\documentclass[runningheads]{llncs}
\usepackage{amsmath,amsfonts,bm}

\def\eqref#1{equation~\ref{#1}}
\def\1{\bm{1}}

\DeclareMathAlphabet{\mathsfit}{\encodingdefault}{\sfdefault}{m}{sl}
\SetMathAlphabet{\mathsfit}{bold}{\encodingdefault}{\sfdefault}{bx}{n}

\DeclareMathOperator*{\argmax}{arg\,max}

\usepackage{hyperref}
\usepackage{url}
\usepackage{graphicx}
\usepackage{arydshln} 
\usepackage{amssymb}

\usepackage[misc]{ifsym}

\usepackage{algorithm2e}
\RestyleAlgo{ruled}
\SetKwComment{Comment}{/* }{ */}

\begin{document}
\title{Learning Objective-Specific Active Learning Strategies with Attentive Neural Processes}
% \title{Contribution Title\thanks{Supported by organization x.}}
%
\titlerunning{Learning Objective-Specific Active Learning}

\author{Tim Bakker\inst{1}(\Letter) \and
Herke van Hoof\inst{1} \and
Max Welling\inst{1,2,3,4}}

\authorrunning{T. Bakker et al.}

\toctitle{Learning Objective-Specific Active Learning Strategies with Attentive Neural Processes}
\tocauthor{Tim~Bakker, Herke~van~Hoof, Max~Welling}

\institute{AMLab, University of Amsterdam (UvA) \\
\email{\{t.b.bakker, h.c.vanhoof, m.welling\}@uva.nl} \and
Canadian Institute for Advanced Research (CIFAR) \and
European Lab for Learning and Intelligent Systems (ELLIS) \and
Microsoft Research (MSR)
}

\maketitle              % typeset the header of the contribution

\begin{abstract}
Pool-based active learning (AL) is a promising technology for increasing data-efficiency of machine learning models. However, surveys show that performance of recent AL methods is very sensitive to the choice of dataset and training setting, making them unsuitable for general application. In order to tackle this problem, the field Learning Active Learning (LAL) suggests to learn the active learning strategy itself, allowing it to adapt to the given setting. In this work, we propose a novel LAL method for classification that exploits symmetry and independence properties of the active learning problem with an Attentive Conditional Neural Process model. Our approach is based on learning from a myopic oracle, which gives our model the ability to adapt to non-standard objectives, such as those that do not equally weight the error on all data points. We experimentally verify that our Neural Process model outperforms a variety of baselines in these settings. Finally, our experiments show that our model exhibits a tendency towards improved stability to changing datasets. However, performance is sensitive to choice of classifier and more work is necessary to reduce the performance the gap with the myopic oracle and to improve scalability. We present our work as a proof-of-concept for LAL on nonstandard objectives and hope our analysis and modelling considerations inspire future LAL work.

\end{abstract}

\section{Introduction}
Supervised machine learning models rely on large amounts of representative annotated data and the cost of gathering sufficient data can quickly become prohibitive. Active learning (AL) attempts to mitigate this problem through clever selection of data points to be annotated, thereby reducing total data requirements. To achieve this, AL exploits available information about the dataset and/or supervised task model (e.g. an image classifier) to select data points whose labels are expected to lead to the greatest increase in task model performance. Most classical AL strategies are hand-designed heuristics, based on researcher intuition or theoretical arguments \cite{settles}. Recently, much work has been focused on scaling AL to deep learning (DL) settings, which are even more data-hungry \cite{dal2020survey}. Such works for instance combine heuristics with representations learned by neural networks \cite{discral2019,vaal2019,graph2021,graph2021vis}), focus specifically on batch acquisition \cite{coreset2018,bayes2019sparse,kddp2019,badge2020,wasser2020}, or adapt Bayesian Active Learning by Disagreement (BALD) \cite{bald2011,bald2017,baba2021,epigbald2021,bamld2021,powerbald2021,causalbald2021}. Despite these developments, it has been observed that modern AL strategies can vary wildly in performance depending on data setting and that there is no single strategy that consistently performs best \cite{onlinechoice2004,ralf2012,alemp2017survey,dal2020survey,aldata2020survey,aleval2020survey}. This observation has spurred the development of Learning Active Learning (LAL) methods, which attempt to directly learn an active learning strategy on some data. The goal is to either learn a method that is specifically adapted to the data setting at hand \cite{lal2015orig,lal2017,lal2019bnnpg}, or to learn a strategy that performs well for various data settings \cite{lal2018metarl,lal2018uci,lal2018imit,lal2021imital}. Such methods have the potential of adapting to additional properties of the task as well, such as nonstandard objectives. A prominent real-life example of such objectives appears in imbalanced data settings, where rare classes are typically more important than their standard contribution to the loss or accuracy suggests. Current active learning surveys generally focus on balanced data settings; few large-scale empirical studies exist for alternative objectives, such as imbalanced data and AL methods designed to work with imbalanced data. In this paper, we propose a novel Learning Active Learning (LAL) method for pool-based active learning. The model learns from a myopic oracle, which gives it the ability to adapt to objectives besides standard classification accuracy. We validate our model in imbalanced data settings, where we show that 1) existing AL methods underperform, and 2) the myopic oracle provides a strong signal for learning. Our contributions are as follows:\footnote{Experiment code can be found at: \url{https://github.com/Timsey/npal}.}
\begin{enumerate}
    \item We show that a wide range of current pool-based AL methods do not outperform uniformly random acquisition on average across multiple deep learning image classification benchmarks. The tested methods generally perform worse on imbalanced data settings than on balanced data settings, suggesting that current AL methods may be under-optimised for the former.
    \item We present experiments with a myopic oracle that show large performance gains over standard AL methods on simple benchmarks. We observe that these gains are larger for imbalanced data settings, suggesting the oracle exploits specific highly-informative samples during acquisition.
    \item We propose a novel LAL method based on Attentive Conditional Neural Processes that learn from the myopic oracle. The model naturally exploits symmetries and independence properties of the active learning problem. In contrast to many existing LAL methods, it is not restricted to heuristics and requires no additional data and/or feature engineering.
\end{enumerate}

\section{Related Work}
The field of active learning has a rich history going back decades, with the current taxonomy of methods founded on the extensive survey by \cite{settles}. In this work, we focus on \textit{pool-based} active learning, where a `pool' of unlabelled data points is available, and the goal is to select one or more of these to label (i.e. `acquire' the label). Here we focus on some relevant works, and refer to the supplementary material\footnote{Supplementary material can be found at: \url{https://github.com/Timsey/npal/blob/main/full_paper.pdf}}, for additional discussion.

The aforementioned survey discusses a number of classical pool-based active learning methods, the most notable among which is Uncertainty Sampling. Here label acquisition is determined by the uncertainty of the classifier. How this uncertainty is measured determines the flavour of Uncertainty Sampling: Entropy selects the points that have maximum predictive entropy, Least Confident acquires the sample on which the task model is least confident in its prediction, and Margin selects the data point with the smallest difference in predicted probability for the first and second most likely class. CoreSet \cite{coreset2018} instead take a fully geometric approach to active learning by formulating it as a Core-Set selection problem. Acquisition proceeds through optimising annotated data coverage in some representation space. The authors provide a greedy approximation to their algorithm, called k-Center Greedy, which shows competitive performance while being cheaper to compute. Learning Loss \cite{learningloss2019} adds a loss prediction module to the base task model, motivated by the idea that difficult-to-classify samples are promising acquisition candidates. This module has the goal of predicting the task model's loss on any given data point and is jointly trained with the task model. Unlabelled samples with the highest predicted loss are then acquired after training. 

One potential goal in doing active learning is to select an annotated dataset that represents the true data distribution as well as possible. Based on this idea, Discriminative Active Learning (DAL) \cite{discral2019} learns a classifier (discriminator) to distinguish labelled and unlabelled data based on a representation learned by the task model. Acquisition proceeds by annotating the points that the classifier predicts are most likely to be part of the current unlabelled data pool. Variational Adversarial Active Learning (VAAL) \cite{vaal2019} builds on this idea by setting up a two-play mini-max game where a Discriminator network classifies data points as belonging to the labelled or unlabelled set, based on a representation learned by a Variational AutoEncoder (VAE). The VAE is incentivised to fool the discriminator, such that the resulting discriminator probabilities encode similarity between any data point and the currently annotated set. Acquisition then occurs by choosing the least similar points. \cite{graph2021} is a recent Convolutional Graph Neural Network (GCN) method that represents data points as nodes in a graph instead. It too is trained to distinguish labelled and unlabelled datapoints; after training the point with the highest uncertainty according to the GCN is selected for labelling. By representing the full dataset as a graph, this method can encode relevant correlations between data points explicitly. \cite{graph2021vis} extend this method by using Visual Transformers to learn the graph representation. Although research into active learning methods continues, it has been widely observed that AL strategies performance varies heavily depending on data setting and that there is no single strategy that consistently performs best. Such studies typically focus on balanced data settings \cite{onlinechoice2004,ralf2012,alemp2017survey,dal2020survey,aldata2020survey,aleval2020survey}.

\paragraph{Active Learning for Imbalanced Data:} Compared to the wealth of research on active learning, little work has been done on AL for imbalanced datasets specifically. This further motivates imbalanced data settings as relevant nonstandard objectives for active learning. Existing work in this area typically incorporates explicit class-balancing strategies or additional exploration towards difficult examples. Hybrid Active Learning (HAL) \cite{hal2020} is built on the idea that rare samples may differentiate themselves in feature space. HAL trades off geometry-based exploration (e.g. some average distance to the currently annotated data) with informativeness-based exploitation (e.g. as in Uncertainty Sampling). Class-Balanced Active Learning (CBAL) \cite{cbal2022} combines entropy sampling with a regulariser that assigns high values to rare points. This regulariser is the difference between a desired class-histogram (i.e. fully balanced classes) and the sum of softmax values of currently sampled points. This intuitively will have the effect of selecting rare points more often. \cite{vabal2020} derives an active learning strategy based on selecting the example with the highest estimated probability of misclassification through Bayes' theorem and various approximate distributions learned by VAE. \cite{twostep2020} describe a two-step approach that uses the data's class imbalance profile to switch from classical AL to a class-balancing acquisition function that favours pool points close (in embedding space) to the rarest class in the annotated data. \cite{imbens2018} suggests that doing active learning using the variation ratio of a model ensemble may help counteract imbalance in the data.

\paragraph{Learning Active Learning:} With the observation that existing AL methods do not consistently perform well across data settings, interest in learning pool-based active learning has risen. The seminal paper by \cite{lal2015orig} formulates Active Learning By Learning (ALBL) as a multi-armed bandit problem, where the arms are different AL heuristics. The goal is to learn to select the best heuristic for each acquisition round. \cite{lal2019bnnpg} learns to fine-tune existing AL heuristics using a Bayesian acquisition net trained with the REINFORCE algorithm. \cite{lal2018imit} instead learn to imitate actions performed by an approximate oracle. Relatedly, \cite{lal2021imital} reduce the imitation learning goal to a learning-to-rank problem. They meta-train on synthetic data and show this generalises to other datasets. \cite{lal2017} formulates learning active learning as a regression problem. Similarly to our proposed method, they train a model to predict the reduction in generalisation error expected upon adding a label to the dataset. However, their method requires handcrafted global features representing the classification state and annotated dataset as input to their regressor. In contrast, our method implicitly learns the required features from the raw data, allowing for more complex relationships and simplifying engineering choices. Finally, both \cite{lal2018metarl} and \cite{lal2018uci} perform meta-learning over various binary classification datasets. The former employs a meta-network that encodes dataset and classifier states into parameters for a policy, which is reinforcement learned by the REINFORCE algorithm. The latter employs reinforcement learning with a Deep Q-Network and eschews the meta-network. These methods are either still restricted to heuristics \cite{lal2015orig,lal2019bnnpg}, or require gathering additional representative or synthetic datasets for training \cite{ravi2018meta,lal2018metarl,lal2018uci,lal2018imit,lal2021imital}, as well as dataset-independent features. 

\section{A Study on Existing Active Learning Methods} \label{sec:exAL}

In pool-based active learning, we are given a labeled (classification) dataset $\mathcal{D}_{annot} = \{(\bm{x}_i, \bm{y}_i)\}_{i=0}^M$ of size $M$, where $i$ indexes the data points, $\bm{x}_i \in \mathbb{R}^K$ are feature vectors of size $K$, and $\bm{y}_i \in \{0, 1\}^C$ is a (one-hot) label on $C$ total classes. We are further given an unlabelled dataset $\mathcal{D}_{pool} = \{\bm{x}_j\}_{j=0}^N$ of size $N$ and are tasked with selecting candidates $\bm{x}_j$ from $\mathcal{D}_{pool}$ to annotate: i.e. select the index $j$, obtain the label $\bm{y}_j$, and subsequently add $(\bm{x}_j, \bm{y}_j)$ to $\mathcal{D}_{annot}$. The goal of this procedure is to iteratively improve a task model, e.g. a classifier, trained on the annotated data $\mathcal{D}_{annot}$. Improvement is typically measured by some performance metric, e.g. the accuracy on some test dataset $\mathcal{D}_{test}$. Most existing AL methods depend on combinations of heuristics and representation learning for selecting the index $j$. The implicit expectation is that the selections such heuristics make are also highly performant according to the chosen performance metric. Here we explore whether this assumption holds in modern deep active learning.

\paragraph{Data:} To explore the performance of existing heuristic-based AL strategies, we perform active learning on four standard ten-class image classification benchmark datasets: MNIST \cite{mnist}, FashionMNIST \cite{fashionmnist}, SVHN \cite{svhn}, and CIFAR-10 \cite{cifar}. We use a standard ResNet18 convolutional neural network \cite{resnet} as the base classifier. We consider three objective settings for each benchmark: Balanced, Imbalanced, and Imbalanced weighted. In imbalanced settings, half the classes are undersampled by a factor 10. Evaluation is performed with a balanced accuracy metric, where instances from undersampled classes are upweighted such that all classes have the same importance. Imbalance weighted additionally takes these weights into account during training. This mimics objectives in typical imbalanced data applications, where rare class instances are often considered more important than common ones \cite{imbadl2019survey}. Following \cite{graph2021}, we initialise active learning with an annotated dataset $\mathcal{D}_{annot}$ of 1000 data points that follow the specified class ratios; the remaining point also follow these class ratios and are left as the pool dataset $\mathcal{D}_{pool}$. Every acquisition step we batch annotate 1000 points using the specified AL strategy, for a total of ten steps. After each step, we retrain the classifier from scratch. See supplementary~\ref{app:impdetres} for further implementation details.

\paragraph{AL Strategies:} First, we consider the three classical uncertainty sampling strategies \cite{settles}: \textsc{Entropy}, \textsc{Margin} and \textsc{LstConf} (least-confident). Second, we include the purely geometric approach of \cite{coreset2018}: \textsc{KCGrdy} (K-center greedy). Third, active learning through Learning Loss Module: \textsc{LLoss} \cite{learningloss2019}. Fourth, Variational Adversarial Active Learning \textsc{VAAL} \cite{vaal2019}; a discriminator method based on VAE-learned representations. Fifth, two variations on the same convolutional graph neural network method -- \textsc{UncGCN} and \textsc{CoreGCN} \cite{graph2021} -- that employ a jointly learned discriminator and graph embedding; unlike VAAL, this approach can explicitly model inter-datapoint correlations. Sixth, we employ \textsc{HAL} \cite{hal2020} and \textsc{CBAL} \cite{cbal2022} as baselines specifically developed for active learning in imbalanced data settings. \textsc{HAL} is further split into \textsc{HALUni} and \textsc{HALGau}, depending on the exploration scheme (uniform or Gaussian). Finally, \textsc{Random} is the uniformly random sampling baseline, corresponding to no active learning.
\begin{figure}[ht] 
\begin{center}
\includegraphics[width=\textwidth]{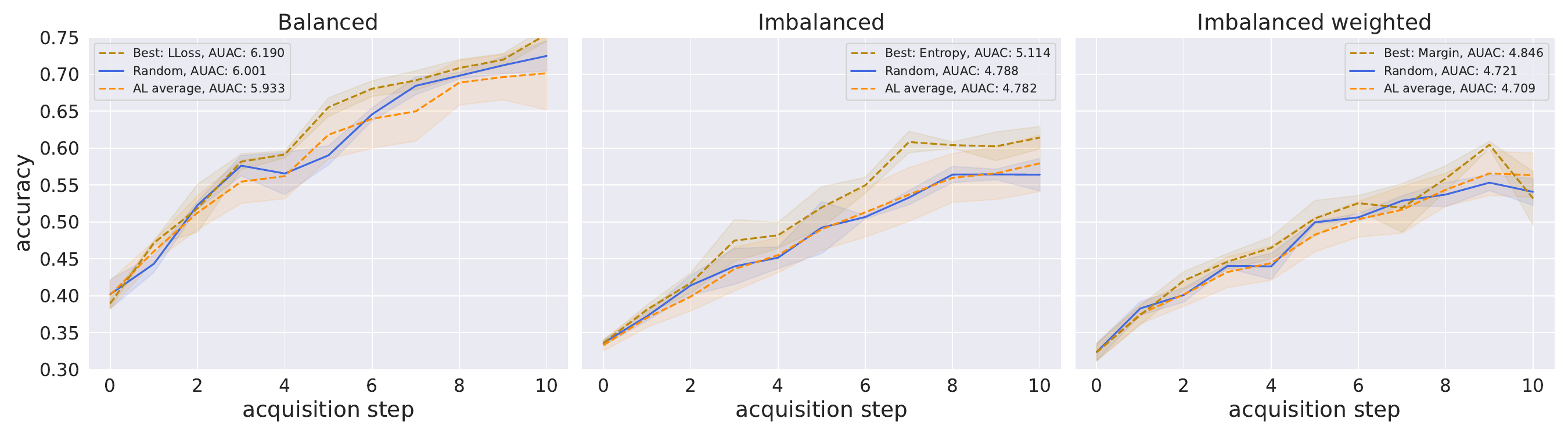}
\end{center}
\caption{Random vs. best and average of remaining AL strategies for CIFAR-10 dataset and ResNet18 classifier, 1000 acquisitions per step, and 1000 initial labels. Shaded region represents standard deviation over three seeds.}
\label{fig:cifar10}
\end{figure}
\paragraph{Results:} In Figure~\ref{fig:cifar10} we plot CIFAR-10 test accuracy as a function of acquisition step for \textsc{Random}, the best performing AL method, and the average of all AL methods (excluding \textsc{Random}). AUAC is the Area Under the Acquisition Curve, which is computed as the area under the curves of Figure~\ref{fig:cifar10}. It measures performance of the whole AL trajectory. We observe that the average active learning strategy does not perform significantly better than \textsc{Random} in any setting. The best performing AL strategy (by AUAC) does outperform \textsc{Random}. These results suggest that AL can be useful, but only if an appropriate strategy is found for the data at hand; a mismatched strategy can lead to performance worse than uniformly random labelling. Note that there is no consistent best performer among the AL methods. This variation in (relative) performance across benchmarks has been previously observed in the literature \cite{onlinechoice2004,ralf2012,alemp2017survey,dal2020survey,aldata2020survey,aleval2020survey}. We refer to the supplementary material for implementation details (\ref{app:impdetres}) and additional results (\ref{app:resnet_exp}). We further argue in \ref{app:resnet_exp} that the tested AL methods generally perform worse on the imbalanced objective settings than on the balanced settings, suggesting that current AL methods may be under-optimised for the former.

\section{Myopic Oracle Active Learning} \label{sec:oracle}

Given the results of the previous section, we may wonder if stronger AL strategies can be found. In particular, it would be valuable to develop strategies that perform well out of the box on many different settings. To this end, the field of Learning Active Learning (LAL) has emerged. The motivating idea is that information about the problem setting should be used for constructing the AL strategy: LAL-methods attempt to do this through learning. What is learned can vary from a choice between existing heuristics \cite{lal2015orig}, to a fine-tuning of such heuristics \cite{lal2019bnnpg}, to a labelling policy that tries to generalise over datasets \cite{lal2018metarl,lal2018uci,lal2018imit,lal2021imital}, to the direct improvement to the underlying classifier upon annotating a data point in the given dataset \cite{lal2017}. Ideally, the learned AL strategy should not be constrained to be close to human heuristics, as there is no guarantee that optimal strategies can be represented as such. Additionally, we will only require the availability of a single dataset to train an AL strategy, since finding additional datasets representative of the problem setting at hand is often not feasible in real-world applications. That leaves us with strategies similar to those in e.g. \cite{lal2017}, where the AL strategy tries to learn a function mapping the features of an unlabelled datapoint to the expected improvement of the classifier after retraining with that datapoint labelled. Before attempting to train such a strategy, we should quantify whether such a method -- if properly learned -- actually improves much over existing methods. To this end, we introduce the myopic oracle strategy -- denoted \textsc{Oracle} in the below -- which computes the actual classifier improvement on the test data for an unlabelled datapoint $\bm{x}_j$ in $\mathcal{D}_{pool}$, by treating the corresponding label $\bm{y}_j$ as known and retraining the classifier with this additional label. This improvement is stored, the classifier is reset, and the process is repeated for every datapoint in $\mathcal{D}_{pool}$. Pseudocode for obtaining improvement scores with the \textsc{Oracle} is presented in supplementary~\ref{app:impdetor}. \textsc{Oracle} then selects the datapoint $(\bm{x}^*, \bm{y}^*)$ corresponding to the largest classifier improvement and this point is added to the annotated dataset $\mathcal{D}_{annot}$. This oracle uses information that is typically unavailable during the AL process, namely the true labels $\bm{y}_j$ and the exact classifier improvements on the test set. The oracle is myopic, as it greedily acquires the best datapoint every acquisition step, rather than planning ahead: looking ahead $t$ acquisition steps requires retraining the classifier $\binom{|\mathcal{D}_{pool}|}{t}$ times, which is infeasible.

\paragraph{Classifiers:} Even for $t=1$, the myopic oracle strategy requires retraining the underlying classifier $|\mathcal{D}_{pool}|$ times every acquisition step, which is computationally intractable for neural network classifiers. For this reason, our experiments in this setting use simpler classification models. We run experiments with logistic regression classifiers and provide additional experiments with support vector machine (SVM) classifiers in the supplementary material (\ref{app:np_exp}). These are both quick-to-train models that have a long history of being used in AL research \cite{settles,svm2007imba,svm2014,aldata2020survey}, including within the subfield of LAL \cite{lal2015orig,lal2017,ravi2018meta,lal2018uci}. For both classifiers, we employ the default scikit-learn implementations \cite{scikit-learn}, with class-weighting when specified.

\paragraph{Data:} These simpler classifiers do not perform well on the image datasets of Section~\ref{sec:exAL}. In order to properly study the effects acquisition has on model performance, we instead use simpler datasets. A popular choice in the field of learning active learning \cite{lal2018uci,ravi2018meta,lal2018imit} are binary classification datasets from the UCI data repository \cite{uci2017}. We use the `waveform', `mushrooms' and `adult' datasets, since these contain sufficient samples for our experiments post-imbalancing. Data is imbalanced by a factor of ten, as in the previous experiments. In all experiments we initialise the runs with 100 annotated examples and acquire one additional label in each of ten acquisition steps. We set aside 200 datapoints as test data $\mathcal{D}_{test}$ for evaluating the classifiers; oracle scores are also computed on this test data.

\paragraph{AL Strategies:} We first compare \textsc{Oracle} with a logistic regression classifier to the same set of AL strategies we compared to in Section~\ref{sec:exAL}. However, we skip the comparisons to \textsc{LLoss}, \textsc{VAAL}, \textsc{UncGCN}, and \textsc{CoreGCN}, since these all require neural network classifiers as their base. Additionally, the three uncertainty sampling methods \textsc{Entropy}, \textsc{Margin}, and \textsc{LstConf} reduce to the same algorithm for binary classification: we henceforth denote this method as \textsc{UncSamp}. Our goal is to work towards a general-purpose AL method that can be trained using only available data. Therefore, we do not include the discussed LAL methods in our baselines, as these methods either adapt existing heuristics or require heavy feature engineering and/or additional datasets to train.

\paragraph{Results:}
\begin{table}[t]
\caption{AL strategy AUAC and final-step test accuracy on UCI waveform dataset with logistic regression classifier, 1 acquisition per step, and 100 initial labels. Averages and standard deviations are computed over nine seeds.}
\label{tab:waveform_logistic}
\begin{center}
\begin{tabular}{p{1.7cm} p{1.7cm} p{1.7cm} p{1.7cm} p{1.7cm} p{1.7cm} p{1.7cm}}
\multicolumn{1}{c}{\bf Strategy}  & \multicolumn{2}{c}{\bf Balanced} & \multicolumn{2}{c}{\bf Imbalanced} & \multicolumn{2}{c}{\bf Imbalanced weighted} \\
& AUAC & Test acc. & AUAC & Test acc. & AUAC & Test acc.
\\ \hline \\
 \textsc{Oracle} & $\mathbf{9.14} \pm 0.12$ & $\mathbf{0.93} \pm 0.01$ & $\mathbf{8.84} \pm 0.39$ & $\mathbf{0.89} \pm 0.04$ & $\mathbf{9.22} \pm 0.18$ & $\mathbf{0.93} \pm 0.02$ \\
\\ \hdashline \\
 \textsc{UncSamp} & $8.67 \pm 0.17$ & $0.87 \pm 0.01$ & $8.40 \pm 0.49$ & $0.85 \pm 0.04$ & $8.55 \pm 0.33$ & $0.86 \pm 0.02$ \\
 \textsc{KCGrdy} & $8.68 \pm 0.28$ & $0.87 \pm 0.03$ & $8.29 \pm 0.49$ & $0.84 \pm 0.04$ & $8.58 \pm 0.37$ & $0.86 \pm 0.03$ \\
 \textsc{HALUni} & $8.66 \pm 0.26$ & $0.87 \pm 0.03$ & $8.11 \pm 0.55$ & $0.81 \pm 0.06$ & $8.45 \pm 0.46$ & $0.85 \pm 0.05$ \\
 \textsc{HALGau} & $8.68 \pm 0.23$ & $0.87 \pm 0.02$ & $8.16 \pm 0.54$ & $0.82 \pm 0.05$ & $8.48 \pm 0.45$ & $0.85 \pm 0.04$ \\
 \textsc{CBAL} & $8.67 \pm 0.15$ & $0.87 \pm 0.02$ & $8.30 \pm 0.45$ & $0.84 \pm 0.04$ & $8.65 \pm 0.34$ & $0.87 \pm 0.03$ \\
 \textsc{Random} & $8.65 \pm 0.23$ & $0.87 \pm 0.02$ & $8.17 \pm 0.54$ & $0.82 \pm 0.05$ & $8.42 \pm 0.48$ & $0.85 \pm 0.05$ \\
 \textsc{NP} & $8.69 \pm 0.19$ & $0.87 \pm 0.02$ & $8.25 \pm 0.53$ & $0.83 \pm 0.05$ & $8.61 \pm 0.33$ & $0.87 \pm 0.03$ \\
\end{tabular}
\end{center}
\end{table}
Table~\ref{tab:waveform_logistic} compares the performance of the \textsc{Oracle} to pre-existing AL methods on the waveform dataset for the logistic regression classifier. The \textsc{NP} method will be introduced and discussed in the next section. It is clear that \textsc{Oracle} dominates all other AL strategies in all settings. Note that AL is only responsible for a small fraction of the total datapoints in the final step here (10 of 110), whereas in the experiments of the previous section, it was responsible for the majority of datapoints (10000 of 11000). As may be observed in the table, such a small number of points is enough to obtain meaningful differences in scores between AL strategies. This indicates that this benchmark contains sufficient variability between strategies to observe meaningful differences in AL quality, making it an appropriate environment for learning active learning. These results suggest that the function represented by \textsc{Oracle} is a strong active learner that adapts to the given objective. Moreover, we note that the performance gap between \textsc{Oracle} and \textsc{Random} -- and more generally between the various AL strategies -- is larger in the imbalanced settings, providing evidence that acquisition choice is more important in these settings; something \textsc{Oracle} can directly exploit. We refer to the supplementary material for implementation details (\ref{app:impdetor}) and additional results (\ref{app:np_exp}). In the next section, we turn our attention to an attempt at learning an approximation to the \textsc{Oracle} using a Neural Process model.

\section{Learning Active Learning with a Neural Process} \label{sec:np}
Our approach will be to learn an approximation to \textsc{Oracle}, by training a model to predict classifier improvement values for every point in $\mathcal{D}_{pool}$, given a context of annotated datapoints and classifier state. However, we cannot train on the true myopic oracle values, as this requires pool data labels and test data that we do not have access to at training time. Instead, we opt to simulate active learning scenarios by subsampling $\mathcal{D}_{annot}$. For these simulated settings we can compute the improvement values that provide the training signal. Our approach will perform the following procedure at every step of the acquisition process:
\begin{enumerate}
    \item Simulate many active learning scenarios by subsampling $\mathcal{D}_{annot}$ into $N_{sim}$ pairs of annotated and pool data $\left( \mathcal{S}_{annot}^{(i)}, \; \mathcal{S}_{pool}^{(i)} \right)$, with $i \in [1, N_{sim}]$.
    \item Use the myopic oracle to compute -- for each point in all the $\mathcal{S}_{pool}^{(i)}$ -- the classifier improvement observed after retraining with that point and its label to the current dataset $\mathcal{S}_{annot}^{(i)}$.
    \item Train a model to predict these improvements from the input $\left( \mathcal{S}_{annot}^{(i)}, \; \mathcal{S}_{pool}^{(i)} \right)$.
\end{enumerate}

The challenge is now to design a model and training setup that can generalise strategies learned in the simulated settings to the full test-time AL setting represented by $\mathcal{D}_{annot}$ and $\mathcal{D}_{pool}$. Here we describe our considerations and resulting approach to this challenge. First, the classifier improvements used for training should not be computed using test data, as this data is not available during training. Instead, we compute these scores on a held-out `reward' dataset $\mathcal{D}_{val}$. In practice, this reward set was used instead of a validation set, so the usual train-val-test split suffices for training our active learner. Second, our problem setup contains permutation symmetries that can be exploited: the (simulated) annotated dataset forms the context that informs the predictions (improvement scores) of our model, but the order of these points does not matter for the prediction: the context representation should be permutation invariant. Additionally, if our model predicts scores for every (simulated) pool datapoint, then these scores should be permutation equivariant: exchanging the index of two pool points should simply exchange the scores.  Third, in the myopic setting, the score of any pool point is independent of any other pool point, so all point points should be treated individually (i.e., not exchange information). This imposes that the model should be invariant to the number of points in $\mathcal{D}_{pool}$. Note that the independence condition is broken in the non-myopic setting, as combinations of pool points can lead to stronger improvements than the individual myopic scores would suggest. The combination of the second and third conditions / inductive biases heavily restrict the choice of model. A natural choice is to use Neural Process (NP) models \cite{np2018,cnp2018,anp2019,dubois2020npf} to learn the approximate \textsc{Oracle}.

\paragraph{The Neural Process:}
\begin{figure}[ht] 
\begin{center}
\includegraphics[width=.80\textwidth]{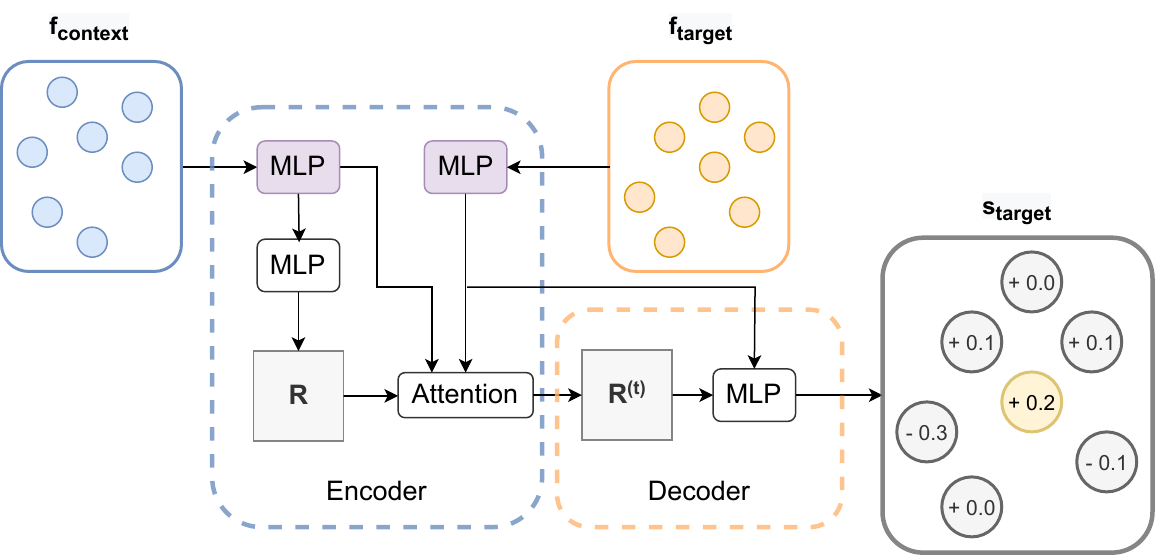}
\end{center}
\caption{Computational graph for the Attentive Conditional Neural Process model. The model takes sets of datapoints as input and predicts improvement values for the target points. Context points correspond to annotated data and target points to pool data. All MLPs are applied pointwise. The top two MLPs (in purple) share weights.}
\label{fig:np_comp}
\end{figure}
The Neural Process comprises a class of models for meta-learning context-conditional predictors and is a natural choice for our approximator. Given a context $\mathcal{C}$ and target input features $\bm{f}_\tau$, the Neural Process outputs a distribution $p(\bm{s}_\tau | \bm{f}_\tau ; \mathcal{C})$ over target predictions $\bm{s}_\tau$. To apply this model to our problem, we identify the context $\mathcal{C}$ with the information stored in the annotated data and the classifier state, the target input features $\bm{f}_\tau$ with the features of pool datapoints, and the target predictions $\bm{s}_\tau$ with the predicted classifier improvements associated to those pool points. We can then train the NP by performing supervised learning -- maximising the log likelihood of target improvements $\bm{y}_\tau$ -- on simulated AL scenarios. At test time we apply the trained model with the full $\mathcal{D}_{annot}$ as context and $\mathcal{D}_{pool}$ as target input. In particular, we utilise an Attentive Conditional NP (AttnCNP) \cite{cnp2018,anp2019}, with cross-attention between the pool and annotated points. The CNP factorises the predictive distribution conditioned on the context set, as
\begin{equation} \label{eq:cnp}
    p(\bm{s}_\tau | \bm{f}_\tau ; \mathcal{C}) = \prod_{t=1}^T p(s^{(t)} | f_\tau^{(t)} ; \mathcal{C} ),
\end{equation}
where $T$ is the number of target datapoints. This modelling choice satisfies the independence of pool point predictions. The context $\mathcal{C}$ should be permutation invariant and is typically encoded into a global representation $R$. The NP is parameterised by a neural network with parameters $\{\theta, \phi\}$ and each factor is typically set to be a Gaussian density \cite{dubois2020npf}, as:
\begin{equation} \label{eq:cnp_full}
    p_{\theta, \phi}(\bm{s}_\tau | \bm{f}_\tau ; \mathcal{C}) = p_{\theta, \phi}(\bm{s}_\tau | \bm{f}_\tau ; R) = \prod_{t=1}^T p_{\theta, \phi}(s^{(t)} | f_\tau^{(t)} ; R ) = \prod_{t=1}^T \mathcal{N} ( s^{(t)} ; \mu^{(t)}, \sigma^{2(t)}),
\end{equation}
where $R = \text{Enc}_\theta (\mathcal{C})$ encodes the context and $(\mu^{(t)}, \sigma^{2(t)}) = \text{Dec}_\phi (R, \bm{f}_\tau^{(t)})$ decodes the context encoding and the target features into target predictive parameters. The AttnCNP extends this model by replacing the global representation $R$ with a target-specific representation $R^{(t)}$ through the use of an attention mechanism. In particular, we use the attention mechanism taken from the Image Transformer \cite{icml2018imtrans} to perform cross-attention between context and target features, constructing $R^{(t)}$. Here context features $\bm{f}_\mathcal{C}$ are treated as keys and target features $\bm{f}_\tau$ as queries. Values are constructed from $\bm{f}_\mathcal{C}$ by applying a pointwise MLP with 2 hidden layers of size 32 and ReLU activations. Our implementation does not use self-attention on the context or target features, as applying self-attention to the target features violates the independence of the pool point scores. In preliminary experimentation, we found that omitting the attention mechanism -- e.g. $R^{(t)} = R$ -- resulted in performance drops due to underfitting the target function, as has been observed in the Neural Process literature \cite{anp2019}. A computational graph of our model is presented in Figure~\ref{fig:np_comp}. This model satisfies the required permutation symmetries while allowing scores of pool points to be given by expressive functions that depend on the context and pool point. In this proof-of-concept study we do not explore the use of uncertainty information for acquisition, rather opting to acquire the datapoint for which $\mu^{(t)}$ -- the predicted mean score -- is maximal, as $j = \argmax_{t \in [1, T]} \mu^{(t)}$. We then acquire the pool datapoint with index $j$, completing a single step in the Active Learning process. The Neural Process is then initialised from scratch, in preparation for the next acquisition step.
% \begin{algorithm}
% \caption{Training the \textsc{NP} model.} \label{alg:np_train}
% \KwData{Annotated dataset $\mathcal{D}_{annot}$, Neural Process model $\textsc{NP}_\theta$ with parameters $\theta$ and fitting function \textsc{.fit(.)}, number of simulations $N_{sim}$, set of fractions $Q$, oracle \textsc{Oracle}, base classifier model $C$, scoring function \textsc{score} evaluated on $\mathcal{D}_{val}$.}
% \KwResult{Trained parameters $\theta^*$ for \textsc{NP}.}
% \For{$i = 1, 2, ..., N_{sim}$}{ 
%     $q_i \gets \text{sample}(Q)$ \Comment*[r]{Uniformly sample an `annotation fraction'}
%     $S_{annot}^{(i)} \gets \varnothing$ \Comment*[r]{Initialise a simulated annotated set}
%     $S_{pool}^{(i)} \gets \mathcal{D}_{annot}$ \Comment*[r]{Initialise a simulated pool set}
%     \While{$|S_{annot}^{(i)}| < round(q_i \cdot |\mathcal{D}_{annot}|)$}{
%         Sample index $j$ of datapoints in $\mathcal{D}_{annot}$ uniformly without replacement \;
%         $S_{annot}^{(i)} \gets S_{annot}^{(i)} \cup (\bm{x}_j, \bm{y}_j)$ \;
%         $S_{pool}^{(i)} \gets S_{pool}^{(i)} \setminus (\bm{x}_j, \bm{y}_j)$ \;
%     }
%     $V_i \gets \textsc{Oracle}\left( S_{annot}^{(i)}, S_{pool}^{(i)}, C, \textsc{score} \right)$ \Comment*[r]{Obtain improvement scores with Algorithm~\ref{alg:oracle} (supplementary material)}
% }

% $\theta^* \gets \textsc{NP}_\theta\textsc{.fit} \left( \{ S_{annot}^{(i)}, \, S_{pool}^{(i)}, \, V_i \}_{i=1}^{N_{sim}} \right)$ \Comment*[r]{Train the \textsc{NP} on the simulated AL settings}

% \Return{$\theta^*$}
% \end{algorithm}

\begin{algorithm}
\caption{Training the \textsc{NP} model.} \label{alg:np_train}
\KwData{Annotated dataset $\mathcal{D}_{annot}$, Neural Process model $\textsc{NP}_\theta$, number of simulations $N_{sim}$, set of fractions $Q$, oracle \textsc{Oracle}, base classifier model $C$, scoring function \textsc{score} evaluated on $\mathcal{D}_{val}$.}
\KwResult{Trained parameters $\theta^*$ for \textsc{NP}.}
\For{$i = 1, 2, ..., N_{sim}$}{ 
    $q_i \gets \text{sample}(Q)$ \Comment*[r]{Uniformly sample an `annotation fraction'}
    $S_{annot}^{(i)} \gets \varnothing$ \Comment*[r]{Initialise a simulated annotated set}
    $S_{pool}^{(i)} \gets \mathcal{D}_{annot}$ \Comment*[r]{Initialise a simulated pool set}
    \While{$|S_{annot}^{(i)}| < round(q_i \cdot |\mathcal{D}_{annot}|)$}{
        Sample index $j$ of datapoints in $\mathcal{D}_{annot}$ uniformly without replacement \;
        $S_{annot}^{(i)} \gets S_{annot}^{(i)} \cup (\bm{x}_j, \bm{y}_j)$ \;
        $S_{pool}^{(i)} \gets S_{pool}^{(i)} \setminus (\bm{x}_j, \bm{y}_j)$ \;
    }
    $V_i \gets \textsc{Oracle}\left( S_{annot}^{(i)}, S_{pool}^{(i)}, C, \textsc{score} \right)$ \Comment*[r]{Obtain improvement scores with Algorithm~\ref{alg:oracle} (supplementary material)}
}

$\theta^* \gets \textsc{NP}_\theta\textsc{.fit} \left( \{ S_{annot}^{(i)}, \, S_{pool}^{(i)}, \, V_i \}_{i=1}^{N_{sim}} \right)$ \Comment*[r]{Train the \textsc{NP} on the simulated AL settings}

\Return{$\theta^*$}
\end{algorithm}

\paragraph{Data:} The experiments for our Neural Process model (\textsc{NP}) are performed on the datasets described in the previous section. In order to train the \textsc{NP} model, we simulate active learning scenarios by sampling from the existing annotated dataset $\mathcal{D}_{annot}$. We define a set of fractions $Q$ and uniformly sample from these a total of $N_{sim}$ times, leading to a set of annotation fractions $\{q_i\}_{i=1}^{N_{sim}}$. For each value of $i$, we then assign the corresponding fraction $q_i$ of datapoints from $\mathcal{D}_{annot}$ to a \textit{simulated} annotated dataset $\mathcal{S}_{annot}^{(i)}$; the remaining points are assigned to a \textit{simulated} pool dataset $\mathcal{S}_{pool}^{(i)}$. This procedure results in a set of $N_{sim}$ simulated/sampled active learning problems of various sizes. We then compute oracle scores of all pool points in each of the resulting AL problems $(\mathcal{S}_{annot}^{(i)}, \mathcal{S}_{pool}^{(i)})$. Since we do not have access to test data at train time, the oracle scores are instead computed on the held-out $\mathcal{D}_{val}$. We present pseudocode in Algorithm~\ref{alg:np_train}. Experimentally we find that simulating with a variety of fractions in $Q$ improves generalisation to the target problem over using a fixed single fraction. Our experiments use $Q = \{0.1, 0.2, ..., 0.8, 0.9\}$ and $N_{sim} = 300$. Preliminary experimentation showed no performance increase for larger values of $N_{sim}$, while using $N_{sim} = 100$ led to slight performance decreases. The held-out dataset $\mathcal{D}_{val}$ consists of the same 100 datapoints for all $i$. 

\paragraph{Results:}
\begin{figure}[ht] 
\begin{center}
\includegraphics[width=\textwidth]{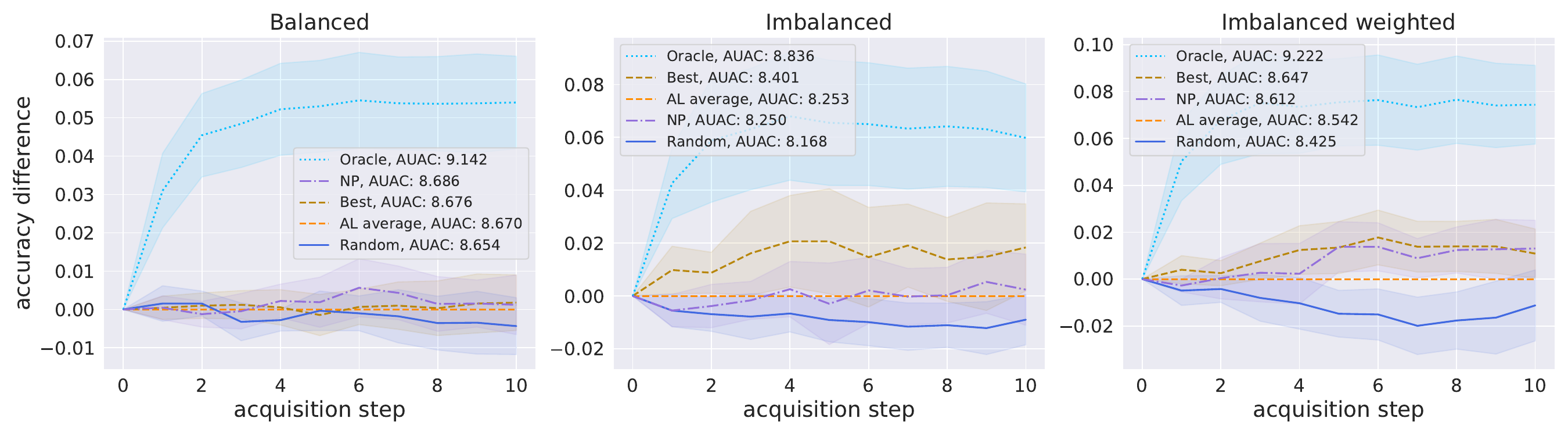}
\end{center}
\caption{Relative performance of acquisition strategies for waveform dataset and logistic regression classifier, 1 acquisition per step, and 100 initial labels. Accuracy differences of \textsc{Random}, \textsc{Oracle}, \textsc{NP} and the best remaining AL strategy (\textsc{Best}) are computed w.r.t. the average of remaining AL strategies (\textsc{AL average}). The shaded region represents twice the standard error of the mean over nine seeds.}
\label{fig:waveform_logistic}
\end{figure}
Table~\ref{tab:waveform_logistic} shows the performance of our method -- \textsc{NP} -- on the UCI waveform dataset with logistic regression classifier. Ignoring \textsc{Oracle}, the Neural Process ranks best of all active learning methods in AUAC on the Balanced setting, second on Imbalanced weighted, and fourth on Imbalanced. In Figure~\ref{fig:waveform_logistic} we show the performance difference between our method and a chosen baseline. Here we choose the average of AL strategies -- \textsc{AL average} -- as the baseline, where we exclude \textsc{Oracle}, \textsc{Random}, and \textsc{NP} from the average. This choice of baseline allows us to clearly see whether any particular method is expected to improve over a naive application of active learning. We also show the performance of the best AL strategy -- \textsc{Best} -- again excluding \textsc{Oracle}, \textsc{Random}, and \textsc{NP} from the selection. This represents the relative performance of choosing the best AL strategy post-hoc. We observe that \textsc{NP} performs on par with \textsc{Best} for Balanced and Imbalanced weighted, and performs similarly to \textsc{AL average} for the Imbalanced setting. In all cases, the gap with \textsc{Oracle} remains large, indicating potential room for improvement. Shaded regions correspond to twice the standard error of the mean, i.e., $2 \cdot \frac{\sigma}{\sqrt{n}}$, where $\sigma$ is the standard deviation and $n$ the number of runs. Tables~\ref{tab:mushrooms_logistic}, \ref{tab:adult_logistic}, and Figures~\ref{fig:mushrooms_logistic}, \ref{fig:adult_logistic} of supplementary~\ref{app:np_exp} similarly show performance on respectively the mushrooms and adult datasets. \textsc{NP} at least slightly outperforms \textsc{AL average} on Imbalanced and Imbalanced weighted for these datasets and in half those cases achieves near-\textsc{Best} performance. However, \textsc{NP} ranks near the bottom in the Balanced setting here. Interestingly, \textsc{Random} outperforms almost all methods on Balanced, possibly indicating increased difficulty in active learning, although \textsc{Oracle} does still demonstrate a large performance gap.

\begin{figure}[ht] 
\begin{center}
\includegraphics[width=\textwidth]{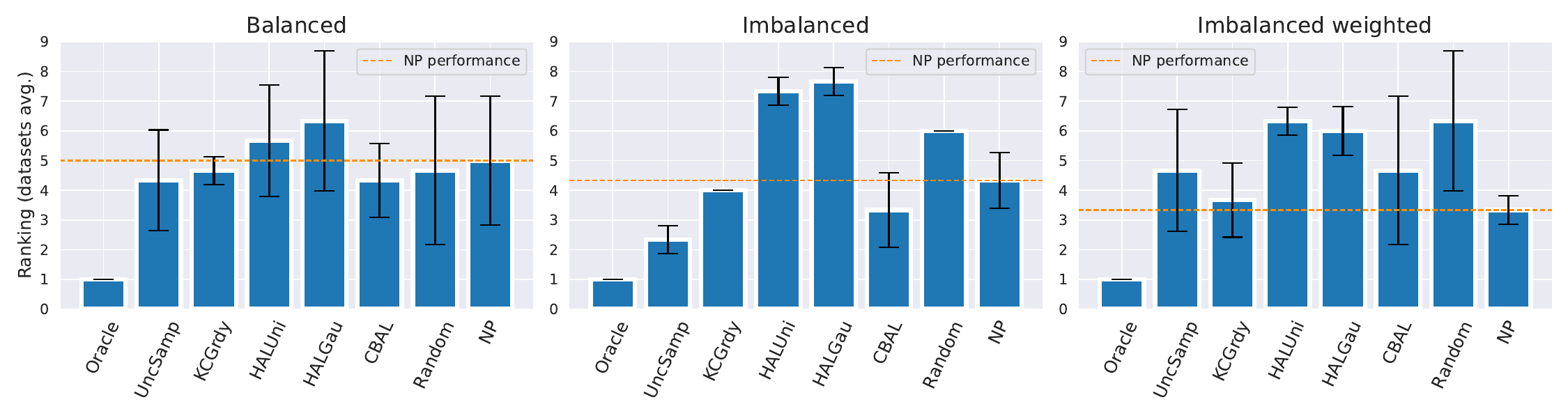}
\end{center}
\caption{Relative AUAC rank of AL strategies averaged over the three UCI datasets for logistic regression. The standard deviation of this rank is denoted by the error bars.}
\label{fig:rank_logistic_AUAC_uci}
\end{figure}
Our method is partially motivated by the need for AL algorithms that perform more stably across different data settings. To this end, Figure~\ref{fig:rank_logistic_AUAC_uci} shows the average AUAC ranking of every AL method across the three UCI datasets. We observe that \textsc{NP} is the best performing AL method on average for the Imbalanced weighted setting and has more middling performance for the other two settings, with Balanced being the worst for our model. Inspecting the ranking standard deviation, we further see that our model achieves a relatively stable ranking across the three datasets in the Imbalanced weighted setting. This stability again degrades for Imbalanced and even further for Balanced. However, note that low standard deviation is only desirable for models with low performance rank, as it otherwise indicates a stable underperformance. These results suggest that \textsc{NP} is better able to exploit information encoded by the \textsc{Oracle} in imbalanced settings. We present results for the SVM classifier in supplementary~\ref{app:np_exp}. This setting seems more difficult for \textsc{NP} to learn, suggesting that the choice of the underlying classifier is important. Figure~\ref{fig:rank_svm_AUAC_uci} corroborates this hypothesis, although it again suggests that \textsc{NP} is specifically more promising for imbalanced settings. 

\section{Conclusion and discussion}
It has been observed in the literature that a wide range of current pool-based active learning methods do not perform better than uniform acquisition on average across standard deep learning benchmarks. We have experimentally verified these results and extended them to imbalanced data settings, which are relevant alternative objectives for many real-world applications. We have explored the validity of using a myopic oracle as a target function for learning active learning (LAL) and have shown its dominating performance on simple active learning tasks. Finally, we have identified symmetry and independence properties of such active learning problems and have modelled these using an Attentive Conditional Neural Process. Unlike existing LAL methods, our model (\textsc{NP}) is not based on existing heuristics, and requires no feature engineering and/or additional datasets to train. Our model generally outperforms the average of the competing AL methods in imbalanced data settings, and occasionally all of them individually. However, future work is needed to evaluate performance on additional datasets, reduce the performance gap with the myopic oracle, and improve scalability. We present our work as a proof-of-concept for LAL on nonstandard objectives -- with a focus on imbalanced data settings -- and hope our analysis and modelling considerations inspire future LAL work.

\paragraph{Limitations:} The primary limitation of our Neural Process approach is scalability. Supervised learning on the myopic oracle requires retraining the base classifier a large number of times during \textsc{NP} training, which is infeasible for large neural network models. Future work may explore to which degree functions learned on simple classifiers can be transferred to more powerful models. Next, the acquisition procedure may be improved through the use of uncertainty information present in the NP. Finally, the NP input may be augmented with additional features -- e.g., predicted class probabilities of pool points -- to potentially improve learning. Preliminary experimentation showed little effect on performance: we leave further exploring the use of extra features for future work. 

\subsubsection*{Acknowledgments.} This work is supported by the ‘Efficient Deep Learning’ (EDL, \url{https://efficientdeeplearning.nl}) research programme, which is financed by the Dutch Research Council (NWO) domain Applied and Engineering Sciences (TTW). We are have used Weights\&Biases \cite{wandb} for experiment tracking.

\newpage

\subsubsection*{Ethical Considerations.} In recent years, machine learning has had a large impact on society by enabling the development of a variety of new, widely-deployed technologies. Opinions on the value of such technologies vary, but it is clear that they have had both positive and negative impacts. Our research topic of active learning is a promising technology for increasing the efficiency of machine learning model training. Developments in active learning may reduce barrier-to-entry for training and deploying high-performing predictive models, which potentially has both positive and negative downstream consequences. On the positive side, wider access to strong models may increase the adoption of life-saving or simply quality-of-life-improving technologies. Additionally, it may allow relatively less powerful interest groups to not fall behind larger or more powerful institutions in capabilities, thus supporting democratisation of AI. On the negative side, improved active learning has the potential to exacerbate the negative effects of machine learning applications as well. Such exacerbation may happen through widening the aforementioned capability gap between less and more powerful institutions (e.g., by potentially easing model scaling), or through reinforcing existing model biases during training. Additionally, training large-scale models consumes a large amount of energy, potentially worsening the current energy and climate crises. Finally, any machine learning capabilities research potentially exacerbates the future risks of AI misalignment; risks that are worrying to an increasing share of the research community \cite{grace2023impacts,cais2023statement}.


\bibliographystyle{splncs04}
\bibliography{references.bib}

\newpage 

% Appendix: not used in submission
\appendix
\section{Additional related work} \label{app:related}
Active learning being such a large research field, it is naturally infeasible to refer to all related work. Nonetheless, we include some additional related research here -- primarily on balanced non-LAL settings -- to give a more comprehensive overview of the literature.

Two more notable classes of active learning strategies discussed in the classical survey \cite{settles} are Query-by-Committee (QBC), and Expected Error Reduction (EER). In QBC, multiple task models are trained on the available labelled data, with each model representing a different valid hypothesis. The acquired point(s) are those on which the models have the strongest disagreement in their predictions, by some metric of disagreement, such as such as vote entropy or Kullback-Leibler (KL) divergence. EER seeks to acquire labels for samples that lead to the biggest reduction in generalisation error, typically estimated as the expected error (under the current task model) of the task model on the remaining unlabelled data points. Bayesian Active Learning by Disagreement (BALD) \cite{bald2011} comprises a class of methods that aim to minimise the task model's expected posterior predictive entropy upon acquisition. First adapted to Deep Learning methods by \cite{bald2017}, this algorithm has spawned many variations in the literature \cite{baba2021,epigbald2021,bamld2021,powerbald2021,causalbald2021}. Another Bayesian approach, motivated by matching the log posterior after acquisition to the full data log posterior, is explored in \cite{bayes2019sparse}. A non-Bayesian distribution matching approach \cite{wasser2020} incorporates a diversity term that minimises Wasserstein distance between the full data distribution and the acquired data at any given AL step. 

Since retraining deep learning models with only a single extra data point is expensive and unlikely to be useful, much DL active learning research has been focused on batch efficiency: selecting batches of data, where individual data points do not strongly overlap in information. This introduces a trade-off between \textit{informativeness} and \textit{diversity} of data selected by AL. Some works explicitly model this trade-off: \cite{kddp2019}, which makes a principled informativeness-diversity trade-off using k-Determinantal Point Processes (k-DPP). BADGE \cite{badge2020} uses hallucinated gradients as an informativeness proxy, and combines it with a k-DPP sampling procedure that incorporates sample diversity. Generative AL \cite{gaal2017,adval2018,bayes2019gen} comprises a class of methods that use GAN-like models to generate informative samples artificially. These may then be annotated directly. These methods suffer from the problem of generating samples that are meaningless to human annotators. As such, unlabelled points from the pool dataset that are closeby to the generated samples according to some metric are often used as alternative acquisition targets. Influence methods \cite{influence2019,influence2021} attempt to use influence functions to estimate changes in the task model that would occur as a result of labelling a particular sample and retraining the task model. This is similar to our method in that influence-based AL can be used to directly estimate model improvements on any arbitrary differentiable utility function. However, these estimations are based on local linearisations of the loss surface, rather than trained on observed model improvements. Moreover, brief experimentation suggested their performance is highly dependent on finding a stable initialisation for the influence estimation.

\paragraph{Imbalanced Machine Learning:} There is a large literature on solutions to data imbalance which do not involve active learning. \cite{imbadl2019survey} discuss a variety of such methods published on deep learning settings between 2015 and 2018. More recent works cover a wide variety of topics, such as loss function learning for long-tailed data \cite{zhang2017rangeloss}, person re-identification \cite{gomez2019reident}, transfer learning for open-world settings \cite{liu2019openworld}, meta-learning class weights \cite{shu2019metaweightnet}, the value of imbalanced labels \cite{yang2020labelvalue}, imbalanced regression \cite{yang2021imbareg}, semi-supervised learning with minority classes \cite{yuille2021crest}, and self-supervised contrastive learning for long-tailed data \cite{li2022targeted}. Of special interest is \cite{oh2010activeimba}, which performs a form of active pruning: selecting examples from an existing pool of labelled data to train an ensemble of classifiers. Note that this differs from active learning scenarios in that it requires all data to be already labelled. 
\section{Implementation details} \label{app:impdet}
This Supplementary material contains implementation details and additional results for all our experiments. Section~\ref{app:impdetres} presents ResNet18 experiments, section~\ref{app:impdetor} presents Myopic Oracle experiments, and section~\ref{app:impdetnp} presents Neural Process experiments.

\subsection{ResNet experiments} \label{app:impdetres}
In this section, we present some additional implementation details for the ResNet experiments of Section~\ref{sec:exAL}. 

\subsubsection{Classifier.} We employ a standard ResNet18 \cite{resnet} implementation for our classifier model, taken from \cite{graph2021}. The classifier is trained on batches of 128 samples for 200 epochs, with a learning rate of 0.1, momentum 0.9, and weight decay 0.0005. The learning rate is decreased by a factor 10 after 160 epochs. Class-weighted training is performed by adding the relative class weights to the training loss of the classifier.

\subsubsection{Data.} Here we provide additional information on the three data settings: Balanced, Imbalanced, and Imbalanced weighted. First, we consider the balanced case, where every class is represented equally in the training and test data. If the dataset is not naturally balanced, overrepresented class examples are randomly removed until balance is reached. Secondly, we consider a setting in which instances of every even-numbered class (zero-indexed) are ten times undersampled compared to their odd-numbered class counterparts, in both the train and test datasets. For computing test accuracy, we weigh the test accuracy values of every data point by the inverse relative class frequency, such that the initial importance of every class is equal, even though the rare classes contain many fewer data points. This mimics objectives in typical imbalanced data applications, where rare class instances are often considered more important than common ones \cite{imbadl2019survey}. For example, when training a classifier to classify defects as part of a quality-control pipeline for industrial processes, rare defects often lead to more catastrophic failures and so are much more important to detect and classify correctly. Third, we consider the same setting, but now we additionally scale the train loss values of every data point the same way, such that our classifier model is aware of these relative weightings. As a practical and frequently used way of addressing class imbalance, this is an especially relevant nonstandard objective.
\begin{table}[t]
\caption{Dataset information for the ResNet18 experiments.}
\label{tab:data_resnet}
\begin{center}
\begin{tabular}{lcccc}
dataset & \# pool (balanced) & \# pool (imbalanced) & \# features & \# classes
\\ \hline \\
MNIST & 53210 & 28815 & $28 \times 28$ & 10 \\
FashionMNIST & 59000 & 32000 & $28 \times 28$ & 10 \\
SVHN & 45590 & 24620 & $32 \times 32$ & 10 \\
CIFAR-10 & 49000 & 26500 & $32 \times 32$ & 10 \\
\end{tabular}
\end{center}
\end{table}
Dataset details for these setting are presented in Table~\ref{tab:data_resnet}. Each benchmark dataset provides a train and test partition, to which we apply the balancing or imbalancing described above. The image classification datasets come with a pre-determined train-test split, which we use in all our experiments. Due to the imbalancing step, the test data changes between the balanced and imbalanced settings (i.e., there are fewer examples of the rare classes, mimicking the train split). Due to the upweighting of rare classes during evaluation, the expected test error is similar in both settings. However, direct comparisons of the test error between settings should be treated carefully, since the datasets are not exactly equal.

\subsubsection{AL strategies.} Here we provide implementation details for our active learning strategies. Some of the implementations were adapted directly from \cite{graph2021}, which subsamples the pool dataset to 10000 datapoints before running the acquisition strategy, to save on computation. For those methods adapted from \cite{graph2021} we use this subsampling implementation, as it did not affect results significantly compared to full sampling in preliminary experiments.

\paragraph{Uncertainty Sampling:} We use standard implementations of the uncertainty sampling methods. These methods have no hyperparameters beyond the choice of strategy (Entropy, Margin, Least Confident).

\paragraph{HAL:} For both HALUniform and HALGauss, the exploration probability is set to $0.5$. Gaussian exploration in HALGauss is performed with a $\delta$ (which is related to the variance of the Normal) of 10. 

\paragraph{CBAL:} The regularisation hyperparameter $\lambda$ is set to $1.0$ in our experiments.

\paragraph{k-Center Greedy:} Distances are computed using a Euclidean metric on the ResNet feature representation. This representation is the flattened activations of the ResNet before the final Linear layer. The basic implementation was taken from \cite{graph2021}\footnote{\url{https://github.com/razvancaramalau/Sequential-GCN-for-Active-Learning}}.

\paragraph{Learning Loss:} The loss prediction module takes as input the output of the four basic ResNet blocks and is trained using SGD for 200 epochs with learning rate $0.1$, momentum $0.9$ and weight decay $0.0005$. Learning rate is decayed by a factor $10$ after 160 epochs. Margin $\xi$ and loss weight $\lambda$ are both set to $1.0$. Training occurs end-to-end with the ResNet, but uses a different optimiser (SGD vs. Adam). Gradients of the ResNet features feeding into the loss prediction module are detached starting at epoch $120$, to increase stability. The basic implementation was taken from \cite{graph2021}.

\paragraph{VAAL:} VAAL is trained using batch size 128 for 100 epochs with the Adam optimiser and learning rate $0.0005$. It consists of a convolutional VAE with four encoding and decoding layers, and a three-layer Multi-Layer Perceptron with hidden dimension 512 as the Discriminator. Two VAE steps with $\beta = 1.0$ are performed for each batch for every Discriminator step. The loss parameters $\lambda_1$ and $\lambda_2$ are both set to $1.0$, such that the VAE and Discriminator loss are weighted equally. The basic implementation was taken from \cite{graph2021}.

\paragraph{GCN:} The graph network consists of two Graph Convolution layers with hidden dimension 128. Dropout with probability $0.3$ is applied after the first layer. The first layer has ReLU activations, the second has Sigmoid activations that map the feature representation into the two classes: `labelled' and `unlabelled'. The network is trained for 200 epochs by the Adam optimiser with a learning rate of $0.001$ and weight decay $0.0005$. The loss hyperparameter $\lambda$ is set to $1.2$. For UncertainGCN, the margin $s_{margin}$ has been set to $0.1$. For CoreGCN, the k-Center Greedy implementation described above is applied to the graph feature representation. The basic implementation was taken from \cite{graph2021}.

\subsection{Myopic oracle experiments} \label{app:impdetor}
\begin{algorithm}
\caption{Obtaining improvement scores with the \textsc{Oracle}.} \label{alg:oracle}
\KwData{Annotated dataset $\mathcal{D}_{annot}$, pool dataset $\mathcal{D}_{annot}$, base classifier model $C$ with fitting method \textsc{fit(.)}, scoring function \textsc{score} (evaluated on e.g. $\mathcal{D}_{test}$ or $\mathcal{D}_{val}$).}
\KwResult{List $V$ of improvement scores.}
$V \gets \text{empty list}$ \;
$v \gets \textsc{score} \left( C\textsc{.fit} \left( D_{annot} \right) \right)$ \Comment*[r]{Score classifier on simulated data}
\For{$(\bm{x}, \bm{y}) \in D_{pool}$}{
    $v' \gets \textsc{score} \left( C\textsc{.fit} \left( D_{annot} \cup (\bm{x}, \bm{y}) \right) \right)$ \Comment*[r]{Score classifier after adding pool point}
    append $(v' - v)$ to $V$ \;
}
\Return{V}
\end{algorithm}
In this section, we present some additional implementation details for the \textsc{Oracle} experiments of Section~\ref{sec:oracle}. Pseudocode for obtaining improvement scores with the \textsc{Oracle} is presented in Algorithm~\ref{alg:oracle}.

\subsubsection{Data.} The UCI binary classifications datasets used are taken from the code repository\footnote{\url{https://github.com/ksenia-konyushkova/LAL-RL}} corresponding to \cite{lal2018uci}. We selected the three datasets `waveform', `mushrooms' and `adult', as they are still large enough for our experiments after imbalancing.

\begin{table}[t]
\caption{Dataset information for the myopic oracle and Neural Process experiments.}
\label{tab:data_oracle}
\begin{center}
\begin{tabular}{lcccc}
dataset & \# pool (balanced) & \# pool (imbalanced) & \# features & \# classes
\\ \hline \\
waveform & 2894 & 1411 & 21 & 2 \\
mushrooms & 7432 & 3907 & 22 & 2 \\
adult & 390 & 34 & 119 & 2 \\
MNIST & 54010 & 29615 & 728 & 10 \\
\end{tabular}
\end{center}
\end{table}
Table~\ref{tab:data_oracle} shows dataset details for both the balanced and imbalanced settings. These datasets do not come pre-split into train and test data. We split off 200 points as test data and an additional 100 points as reward data for the $\textsc{NP}$ experiments. These splits are controlled by a `data seed' that we vary during our experiments to prevent bias due to dataset selection. For each of these data seeds, we further run multiple experiments with an additional random seed controlling all other randomness in our experiments. In total, we run experiments for three different data seeds, each with three different seeds, for a total of nine experiments per setting. The use of various data splits in our experiments is expected to increase global performance variance, due to larger differences in initialisation, with some data seeds being inherently more challenging than others. We see this effect on the standard deviation in the \textsc{Oracle} and \textsc{NP} experiments. Accuracy differences for e.g. Figure~\ref{fig:waveform_logistic} were computed per seed, to account for the changing train/test data distribution. We additionally run \textsc{Oracle} experiments on the MNIST dataset: these experiments are only performed using the SVM classifier, as logistic regression often failed to converge within the default number of iterations.

\subsubsection{Classifiers.} Most of the AL strategies used in our previous experiments -- all except \textsc{KCGreedy} -- are incompatible with SVM classifiers, as SVMs do not output probabilistic predictions. One could perform Platt scaling \cite{platt99} to turn SVM predictions into probabilities. However, we opt to instead compare to \textsc{FScore}: a method specific to SVM classifiers that acquires datapoints closest to the class-separating hyperplane. This method has shown strong performance in binary classification settings for both balanced and imbalanced settings \cite{svm2007imba}. It is motivated as the strategy that attempts to most rapidly reduce the version space; the space of hypotheses consistent with the observed data \cite{svm2014}. We use the default scikit-learn implementations \cite{scikit-learn} of logistic regression and Support Vector Machine classifiers in our experiments. Class-weighted training is performed using the `sample weights' parameter in the built-in fit function.

\subsubsection{AL strategies.} Here we present additional implementation details for the AL strategies used in our Myopic Oracle experiments of Section~\ref{sec:oracle}. The implementations for Uncertainty Sampling, HAL and CBAL are identical to those described in section~\ref{app:impdetres}. k-Center Greedy is implemented similarly as well, but the greedy min-max problem is solved directly on the input features, rather than on a transformed feature space.

\paragraph{F-Score:} This strategy consists of choosing the pool point with the shortest absolute distance to the separating hyperplane of the trained SVM classifier. This approach is known to have strong performance in the binary classification setting \cite{svm2007imba}. The generalisation to the multi-class setting is not unique: for our MNIST experiments, we choose the point that has the minimum distance to any of the one-versus-rest classification boundaries.

\paragraph{Myopic Oracle:} The myopic oracle chooses the point that maximises test accuracy after retraining, by retraining the underlying classifier on every potential added pool point. This method has no hyperparameters.

\subsection{Neural process experiments} \label{app:impdetnp}
In this section, we present some additional implementation details for the Neural Process experiments of Section~\ref{sec:np}.

\subsubsection{Neural Process model.} The Neural Process implementation used in our experiments is based on the code by \cite{dubois2020npf}\footnote{\url{https://github.com/YannDubs/Neural-Process-Family}}. Our model consists of an Encoder and a Decoder module. The first part of the Encoder is a 1-hidden layer ReLU MLP that is weight-shared between context and target points. This MLP is applied per datapoint to either the context features $\bm{f}_\mathcal{C}$ or the target features $\bm{f}_\tau$, resulting in a datapoint-wise encoding of hidden dimension 32. The context encoding is further processed by the second part of the Encoder -- a 2-layer ReLU MLP -- and combined with the target encoding through an attention mechanism taken from the Image Transformer \cite{icml2018imtrans}. The Decoder takes the resulting representation as input together with the base target encoding and outputs a statistic representing the mean and variance of the prediction for each target datapoint. We only use the mean prediction for our active learning strategy. The model has a total of $21,924$ parameters.

The second part of the Encoder takes context label values as additional input to help predict the target label values. In our implementation, labels represent expected improvement in classification accuracy upon annotating a datapoint, and there are multiple ways to interpret this for context points. The first is to imagine the improvement corresponding to the setting where we have annotations for the entire context except the point in question (a leave-one-out type of setting). This involves computing such improvements on the reward set for the context points to construct the model input. The second is to set this label to 0 for simplicity since we do not expect much model improvement by adding the same data point twice and we already possess an annotation for the data point in question. Preliminary experimentation showed no significant differences between either training method. Our presented results correspond to the second -- simplified -- setting.

One may wonder why we do not use a Latent Neural Process (LNP) \cite{np2018} instead of a CNP, as the former can potentially learn more complex functions. The main draw of the LNP is the ability to learn a joint $p_{\theta}(\bm{s}_{\tau} | \bm{f}_{\tau} ; \mathcal{C})$ that does not factorise conditional on the context $\mathcal{C}$. Due to our assumption of pool point independence, this adds unnecessary complexity. 

\subsubsection{Data and training.} In our experiments the context features $\bm{f}_\mathcal{C}$ and target features $\bm{f}_\tau$  are simply the normalised raw values from the dataset. Additional features -- such as classifier predictions, or context class labels -- can easily be incorporated as well. We observed no significant improvements using target classifier predictions in early experimentation and leave the exploration of including context labels as future work.

The Neural Process model is trained for 100 epochs by the Adam optimiser with a learning rate $0.001$. We exponentially decay the learning rate by a factor 10 during those 100 epochs. The model takes the raw data features -- normalised to the range $[-1, 1]$ -- as input: note that it is possible to add classifier-specific features as well. The target (pool) improvement values are precomputed using the myopic oracle on the reward data $\mathcal{D}_{reward}$.

A single data `point' during training corresponds to a full simulated active learning dataset: i.e., a set of context (annotated) points and a set of target (pool) points, all sampled uniformly from the available annotated dataset $\mathcal{D_{annot}}$. We group simulated AL problems of the same size together, for efficient batching with batch size 64. For sampling the AL problems, we use $Q = \{0.1, 0.2, ..., 0.8, 0.9\}$ and $N_{sim} = 300$. Preliminary experimentation showed no performance increase for larger values of $N_{sim}$, while using $N_{sim} = 100$ led to slight performance decreases. We did not extensively experiment with different values for $Q$. Our simulated AL problems all use the full number of datapoints in $\mathcal{D}_{annot}$ (i.e. no subsampling).

\paragraph{Training Cost Analysis:} A full \textsc{NP} experiment (10 acquisition steps) on the UCI data takes 10-20 minutes on a 1080Ti GPU, with each acquisition step taking under 2 minutes. Training the AttnCNP itself takes only 10s-15s every acquisition step. The bottleneck is in generating the data, due to the repeated retraining of the task classifier. In every acquisition step, this takes 35s-50s for the SVM classifier and 65-85s for the logistic regression classifier. Running an \textsc{NP} experiment for the MNIST dataset with SVM classifier takes a bit under 2 hours on the same hardware. Per acquisition step, dataset creation takes up to 10 minutes (due to the increased number of samples) and NP training about 30s. 

\paragraph{Scalability:} The primary limitation of our Neural Process approach is scalability. Supervised learning on the myopic oracle requires retraining the base classifier a large number of times during \textsc{NP} training, which is infeasible for large neural network models. We note that many potential acquisitions lead to (near-)zero improvement of the classifier: such data points are less interesting for LAL, but take a large fraction of computational resources during training. Strategies for spending less compute on these points may improve scalability.

\section{Additional results} \label{app:exp}

\subsection{ResNet experiments} \label{app:resnet_exp}
In this section, we provide additional results for our ResNet experiments. 

\begin{table}[t]
\caption{AL strategy AUAC and final-step test accuracy on CIFAR-10 dataset with ResNet18 classifier, 1000 acquisitions per step, and 1000 initial labels. Averages and standard deviations are computed over three seeds. The method with the highest mean performance is bolded, as well as any method whose $1$ standard deviation bands include that mean.}
\label{tab:cifar10_resnet}
\begin{center}
\begin{tabular}{p{1.7cm} p{1.7cm} p{1.7cm} p{1.7cm} p{1.7cm} p{1.7cm} p{1.7cm}}
\multicolumn{1}{c}{\bf Strategy}  & \multicolumn{2}{c}{\bf Balanced} & \multicolumn{2}{c}{\bf Imbalanced} & \multicolumn{2}{c}{\bf Imbalanced weighted} \\
& AUAC & Test acc. & AUAC & Test acc. & AUAC & Test acc.
\\ \hline \\
\textsc{Entropy} & $5.99 \pm 0.07$ & $0.69 \pm 0.05$ & $\mathbf{5.11} \pm 0.04$ & $\mathbf{0.61} \pm 0.02$ & $\mathbf{4.84} \pm 0.13$ & $0.58 \pm 0.01$ \\
 \textsc{Margin} & $6.09 \pm 0.03$ & $0.73 \pm 0.00$ & $5.02 \pm 0.07$ & $\mathbf{0.62} \pm 0.02$ & $\mathbf{4.85} \pm 0.09$ & $0.53 \pm 0.04$ \\
 \textsc{LstConf} & $5.92 \pm 0.16$ & $0.74 \pm 0.01$ & $\mathbf{5.07} \pm 0.07$ & $\mathbf{0.63} \pm 0.01$ & $\mathbf{4.80} \pm 0.05$ & $0.57 \pm 0.02$ \\
 \textsc{KCGrdy} & $5.98 \pm 0.02$ & $0.72 \pm 0.02$ & $4.71 \pm 0.04$ & $0.55 \pm 0.01$ & $4.78 \pm 0.02$ & $0.56 \pm 0.02$ \\
 \textsc{LLoss} & $\mathbf{6.19} \pm 0.06$ & $\mathbf{0.75} \pm 0.01$ & $4.85 \pm 0.04$ & $0.61 \pm 0.01$ & $4.75 \pm 0.08$ & $\mathbf{0.61} \pm 0.01$ \\
 \textsc{VAAL} & $6.01 \pm 0.04$ & $0.71 \pm 0.01$ & $4.70 \pm 0.05$ & $0.55 \pm 0.04$ & $4.66 \pm 0.04$ & $0.55 \pm 0.04$ \\
 \textsc{UncGCN} & $6.02 \pm 0.05$ & $0.71 \pm 0.00$ & $4.77 \pm 0.07$ & $0.56 \pm 0.01$ & $4.69 \pm 0.03$ & $0.56 \pm 0.02$ \\
 \textsc{CoreGCN} & $6.05 \pm 0.04$ & $0.70 \pm 0.01$ & $4.75 \pm 0.08$ & $0.59 \pm 0.01$ & $\mathbf{4.78} \pm 0.09$ & $0.59 \pm 0.01$ \\
 \textsc{HALUni} & $5.70 \pm 0.03$ & $0.63 \pm 0.00$ & $4.37 \pm 0.04$ & $0.51 \pm 0.01$ & $4.44 \pm 0.06$ & $0.55 \pm 0.01$ \\
 \textsc{HALGau} & $5.32 \pm 0.04$ & $0.60 \pm 0.04$ & $4.52 \pm 0.08$ & $0.55 \pm 0.01$ & $4.58 \pm 0.13$ & $0.54 \pm 0.02$ \\
 \textsc{CBAL} & $6.02 \pm 0.02$ & $0.72 \pm 0.00$ & $4.73 \pm 0.09$ & $0.59 \pm 0.01$ & $4.63 \pm 0.06$ & $0.55 \pm 0.01$ \\
 \textsc{Random} & $6.00 \pm 0.05$ & $0.72 \pm 0.02$ & $4.79 \pm 0.03$ & $0.56 \pm 0.02$ & $4.72 \pm 0.02$ & $0.54 \pm 0.02$ \\
\end{tabular}
\end{center}
\end{table}

Table~\ref{tab:cifar10_resnet} shows performance on CIFAR-10 per imbalancing setting. Area Under the Acquisition Curve (AUAC) computes the area under the accuracy per acquisition-step curve, such as those depicted in Figure~\ref{fig:cifar10}. This is a measure of performance over the entire acquisition trajectory. We also report the final test accuracy score. As expected, performance is best across the board in the Balanced setting. Interestingly, class-weighted training (Imbalance weighted) seems mostly detrimental to performance, although the effect is less pronounced for the FashionMNIST and SVHN datasets, and reversed for MNIST (see Tables~\ref{tab:svhn_resnet}, \ref{tab:fashion_mnist_resnet}, and~\ref{tab:mnist_resnet}). Note that there is no consistent best performer among the AL methods. While the three uncertainty sampling methods \textsc{Entropy}, \textsc{Margin}, and \textsc{LstConf} are strong performers across datasets, their rank-order varies with dataset, objective setting, and metric. This variation in (relative) performance across benchmarks has been previously observed in the literature \cite{onlinechoice2004,ralf2012,alemp2017survey,dal2020survey,aldata2020survey,aleval2020survey}.

\subsubsection{Comparing active learning across objectives.}
\begin{figure}[ht] 
\begin{center}
\includegraphics[width=\textwidth]{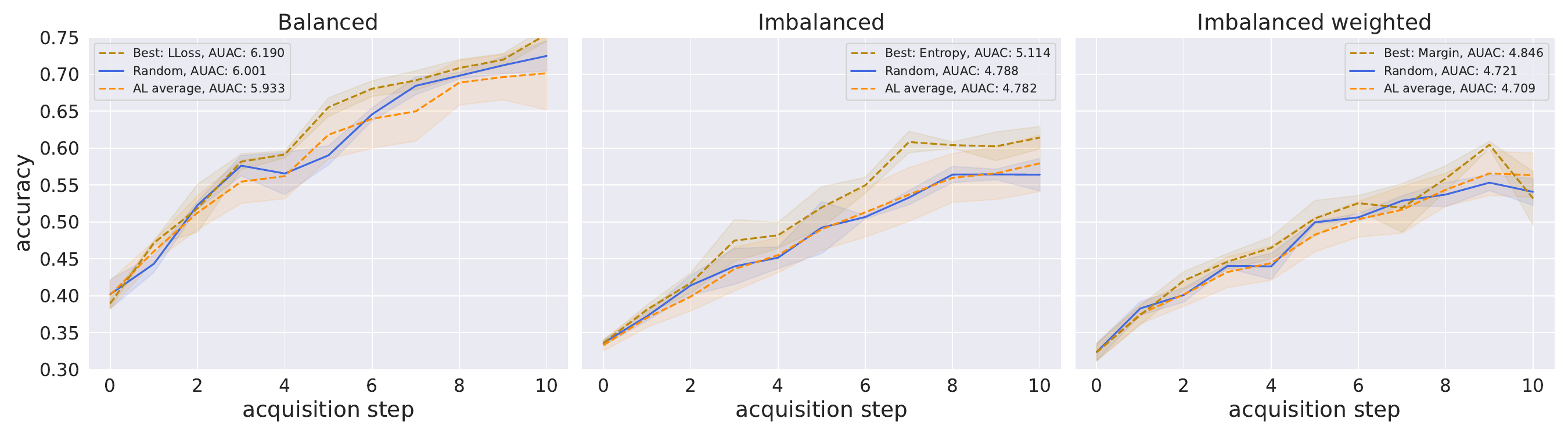}
\end{center}
\caption{Random vs. best and average of remaining AL strategies for CIFAR-10 dataset and ResNet18 classifier, 1000 acquisitions per step, and 1000 initial labels. The shaded region represents the standard deviation over three seeds.}
\label{fig:cifar10app}
\end{figure}

\begin{figure}[ht] 
\begin{center}
\includegraphics[width=\textwidth]{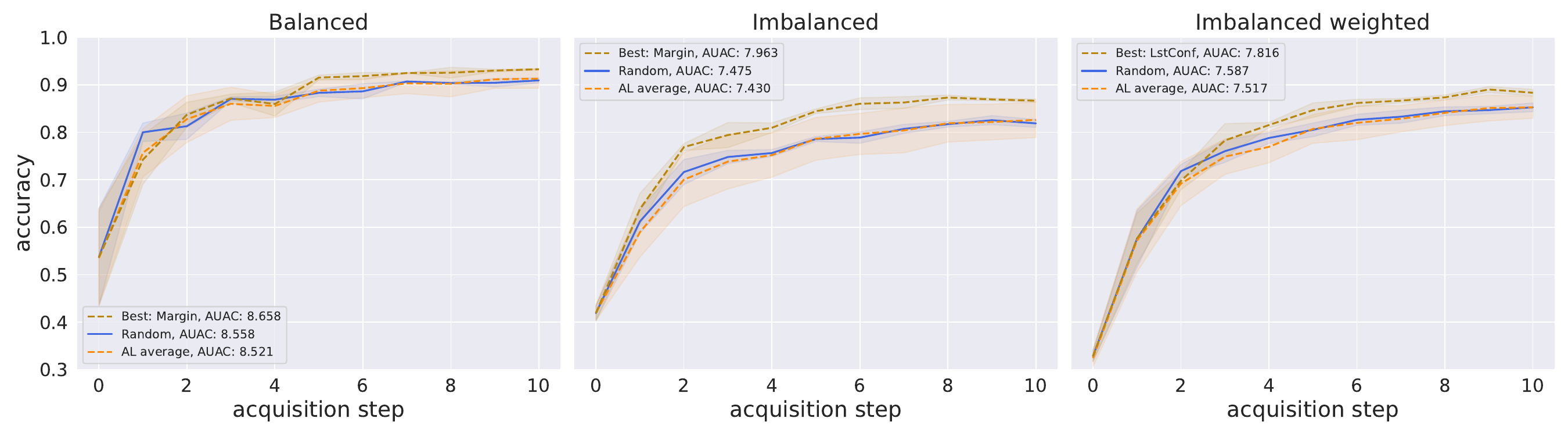}
\end{center}
\caption{Random vs. best and average of remaining AL strategies for SVHN dataset and ResNet18 classifier, 1000 acquisitions per step, and 1000 initial labels.}
\label{fig:svhn}
\end{figure}

\begin{figure}[ht] 
\begin{center}
\includegraphics[width=\textwidth]{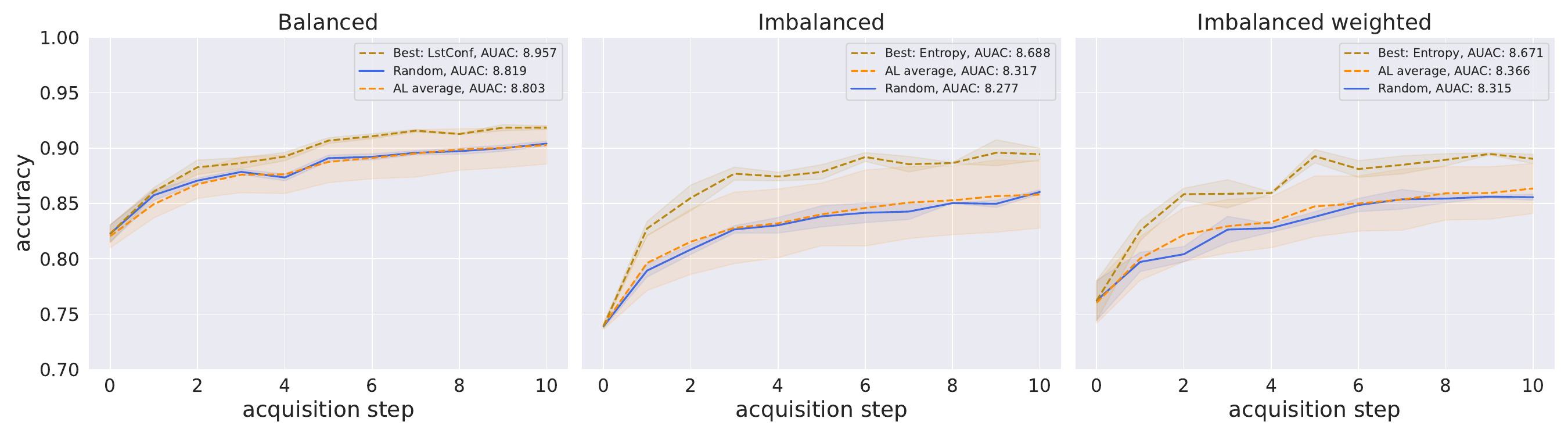}
\end{center}
\caption{Random vs. best and average of remaining AL strategies for FashionMNIST dataset and ResNet18 classifier, 1000 acquisitions per step, and 1000 initial labels.}
\label{fig:fashion_mnist}
\end{figure}

\begin{figure}[ht] 
\begin{center}
\includegraphics[width=\textwidth]{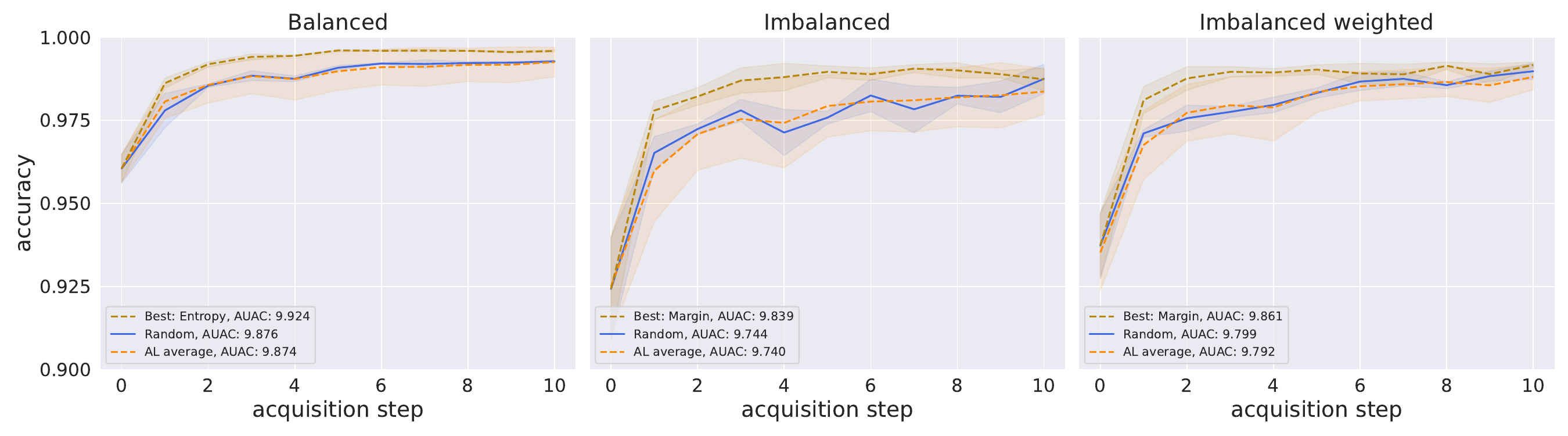}
\end{center}
\caption{Random vs. best and average of remaining AL strategies for MNIST dataset and ResNet18 classifier, 1000 acquisitions per step, and 1000 initial labels.}
\label{fig:mnist}
\end{figure}

In order to compare the performance of active learning across objectives, we once again present Figure~\ref{fig:cifar10} from the main text in Figure~\ref{fig:cifar10app} above. For an ideal strategy, a single acquisition round in the Imbalanced setting should be sufficient to achieve approximately Balanced performance: since we initialise AL with 1000 points and acquire 1000 more every acquisition round, a class-balancing AL strategy could `just' pick mostly rare classes to fully balance the dataset after one step. In practice, this is more difficult, since the labels are unknown. Still, we expect a strong active learner to achieve performance on Imbalanced not much worse than Balanced after a number of such acquisition steps, assuming sufficient rare class examples exist. For the CIFAR-10 experiments, there are 500 examples of each rare class in the training data, so exact balancing is possible until acquisition step four (5000 total datapoints). In Figure~\ref{fig:cifar10}, we observe that the best AL strategy in the Imbalanced setting does not come close to \textsc{Random} performance in the Balanced setting. In fact, the gap in performance widens with the number of acquisition steps, indicating that the tested active learning methods themselves perform worse in imbalanced settings than in balanced settings. This is even true of the methods designed for imbalanced data (\textsc{CBAL}, \textsc{HALUni}, \textsc{HALGau}). However, note that there is a slight variation in test data between objectives due to the imbalancing step, as noted in~\ref{app:impdetres}. As such, direct comparisons between these test accuracies should be treated carefully. For the other three benchmarks, active learning has slightly improved relative performance, as can be seen in Tables~\ref{tab:svhn_resnet}, \ref{tab:fashion_mnist_resnet}, \ref{tab:mnist_resnet}, and Figures~\ref{fig:svhn}, \ref{fig:fashion_mnist}, \ref{fig:mnist}. Still, the average of AL methods does not significantly outperform \textsc{Random} in any setting and even the best strategies in imbalanced settings reach Balanced \textsc{Random} performance in fewer than half the experiments only. In conclusion, it seems that the tested AL methods generally perform worse in imbalanced data settings than in balanced data settings, suggesting that current AL methods may be under-optimised for the former.

\begin{table}[t]
\caption{AL strategy AUAC and final-step test accuracy on SVHN dataset with ResNet18 classifier, 1000 acquisitions per step and 1000 initial labels. Averages and standard deviations are computed over three seeds. The method with the highest mean performance is bolded, as well as any method whose $1$ standard deviation bands include that mean.}
\label{tab:svhn_resnet}
\begin{center}
\begin{tabular}{p{1.7cm} p{1.7cm} p{1.7cm} p{1.7cm} p{1.7cm} p{1.7cm} p{1.7cm}}
\multicolumn{1}{c}{\bf Strategy}  & \multicolumn{2}{c}{\bf Balanced} & \multicolumn{2}{c}{\bf Imbalanced} & \multicolumn{2}{c}{\bf Imbalanced weighted} \\
& AUAC & Test acc. & AUAC & Test acc. & AUAC & Test acc.
\\ \hline \\
 \textsc{Entropy} & $8.47 \pm 0.03$ & $\mathbf{0.94} \pm 0.01$ & $\mathbf{7.91} \pm 0.09$ & $0.86 \pm 0.01$ & $\mathbf{7.72} \pm 0.14$ & $\mathbf{0.89} \pm 0.01$ \\
 \textsc{Margin} & $\mathbf{8.66} \pm 0.07$ & $0.93 \pm 0.00$ & $\mathbf{7.96} \pm 0.04$ & $0.87 \pm 0.00$ & $\mathbf{7.76} \pm 0.11$ & $0.88 \pm 0.01$ \\
 \textsc{LstConf} & $8.62 \pm 0.04$ & $0.93 \pm 0.01$ & $\mathbf{7.93} \pm 0.08$ & $\mathbf{0.87} \pm 0.01$ & $\mathbf{7.82} \pm 0.05$ & $\mathbf{0.88} \pm 0.01$ \\
 \textsc{KCGrdy} & $\mathbf{8.60} \pm 0.08$ & $0.92 \pm 0.01$ & $7.45 \pm 0.03$ & $0.83 \pm 0.01$ & $7.57 \pm 0.12$ & $0.85 \pm 0.00$ \\
 \textsc{LLoss} & $\mathbf{8.64} \pm 0.03$ & $0.92 \pm 0.00$ & $7.32 \pm 0.01$ & $0.82 \pm 0.01$ & $7.29 \pm 0.12$ & $0.86 \pm 0.01$ \\
 \textsc{VAAL} & $8.55 \pm 0.07$ & $0.91 \pm 0.01$ & $7.45 \pm 0.05$ & $0.82 \pm 0.00$ & $7.54 \pm 0.11$ & $0.85 \pm 0.01$ \\
 \textsc{UncGCN} & $8.57 \pm 0.08$ & $0.92 \pm 0.00$ & $7.40 \pm 0.05$ & $0.83 \pm 0.01$ & $7.55 \pm 0.16$ & $0.85 \pm 0.01$ \\
 \textsc{CoreGCN} & $8.58 \pm 0.08$ & $0.92 \pm 0.01$ & $7.31 \pm 0.02$ & $0.82 \pm 0.00$ & $7.50 \pm 0.10$ & $0.84 \pm 0.01$ \\
 \textsc{HALUni} & $8.41 \pm 0.03$ & $0.90 \pm 0.00$ & $7.06 \pm 0.08$ & $0.80 \pm 0.01$ & $7.46 \pm 0.10$ & $0.84 \pm 0.00$ \\
 \textsc{HALGau} & $8.09 \pm 0.06$ & $0.86 \pm 0.01$ & $6.58 \pm 0.10$ & $0.73 \pm 0.02$ & $6.99 \pm 0.09$ & $0.81 \pm 0.01$ \\
 \textsc{CBAL} & $8.54 \pm 0.05$ & $0.91 \pm 0.01$ & $7.35 \pm 0.10$ & $0.83 \pm 0.00$ & $7.50 \pm 0.15$ & $0.84 \pm 0.01$ \\
 \textsc{Random} & $8.56 \pm 0.06$ & $0.91 \pm 0.01$ & $7.48 \pm 0.03$ & $0.82 \pm 0.01$ & $7.59 \pm 0.08$ & $0.85 \pm 0.01$ \\
\end{tabular}
\end{center}
\end{table}

\begin{table}[t]
\caption{AL strategy AUAC and final-step test accuracy on FashionMNIST dataset with ResNet18 classifier, 1000 acquisitions per step, and 1000 initial labels. Averages and standard deviations are computed over three seeds. The method with the highest mean performance is bolded, as well as any method whose $1$ standard deviation bands include that mean.}
\label{tab:fashion_mnist_resnet}
\begin{center}
\begin{tabular}{p{1.7cm} p{1.7cm} p{1.7cm} p{1.7cm} p{1.7cm} p{1.7cm} p{1.7cm}}
\multicolumn{1}{c}{\bf Strategy}  & \multicolumn{2}{c}{\bf Balanced} & \multicolumn{2}{c}{\bf Imbalanced} & \multicolumn{2}{c}{\bf Imbalanced weighted} \\
& AUAC & Test acc. & AUAC & Test acc. & AUAC & Test acc.
\\ \hline \\
 \textsc{Entropy} & $8.94 \pm 0.01$ & $\mathbf{0.92} \pm 0.00$ & $\mathbf{8.69} \pm 0.03$ & $\mathbf{0.89} \pm 0.01$ & $\mathbf{8.67} \pm 0.04$ & $\mathbf{0.89} \pm 0.00$ \\
 \textsc{Margin} & $8.95 \pm 0.01$ & $\mathbf{0.92} \pm 0.00$ & $\mathbf{8.69} \pm 0.04$ & $\mathbf{0.89} \pm 0.01$ & $\mathbf{8.65} \pm 0.03$ & $\mathbf{0.89} \pm 0.01$ \\
 \textsc{LstConf} & $\mathbf{8.96} \pm 0.02$ & $0.92 \pm 0.00$ & $\mathbf{8.67} \pm 0.02$ & $\mathbf{0.89} \pm 0.01$ & $\mathbf{8.64} \pm 0.05$ & $\mathbf{0.89} \pm 0.01$ \\
 \textsc{KCGrdy} & $8.82 \pm 0.01$ & $0.90 \pm 0.00$ & $8.28 \pm 0.01$ & $0.86 \pm 0.01$ & $8.36 \pm 0.05$ & $0.86 \pm 0.01$ \\
 \textsc{LLoss} & $8.79 \pm 0.02$ & $0.90 \pm 0.00$ & $8.28 \pm 0.05$ & $0.86 \pm 0.01$ & $8.27 \pm 0.04$ & $0.86 \pm 0.01$ \\
 \textsc{VAAL} & $8.83 \pm 0.01$ & $0.90 \pm 0.00$ & $8.24 \pm 0.04$ & $0.86 \pm 0.01$ & $8.29 \pm 0.01$ & $0.86 \pm 0.01$ \\
 \textsc{UncGCN} & $8.80 \pm 0.02$ & $0.91 \pm 0.00$ & $8.26 \pm 0.07$ & $0.84 \pm 0.01$ & $8.30 \pm 0.06$ & $0.86 \pm 0.01$ \\
 \textsc{CoreGCN} & $8.83 \pm 0.02$ & $0.90 \pm 0.00$ & $8.29 \pm 0.04$ & $0.87 \pm 0.00$ & $8.38 \pm 0.02$ & $0.86 \pm 0.01$ \\
 \textsc{HALUni} & $8.72 \pm 0.02$ & $0.89 \pm 0.00$ & $8.07 \pm 0.03$ & $0.85 \pm 0.00$ & $8.25 \pm 0.05$ & $0.86 \pm 0.01$ \\
 \textsc{HALGau} & $8.37 \pm 0.04$ & $0.86 \pm 0.01$ & $7.72 \pm 0.06$ & $0.78 \pm 0.01$ & $7.90 \pm 0.06$ & $0.81 \pm 0.01$ \\
 \textsc{CBAL} & $8.83 \pm 0.01$ & $0.90 \pm 0.00$ & $8.30 \pm 0.02$ & $0.85 \pm 0.01$ & $8.31 \pm 0.05$ & $0.86 \pm 0.01$ \\
 \textsc{Random} & $8.82 \pm 0.01$ & $0.90 \pm 0.00$ & $8.28 \pm 0.04$ & $0.86 \pm 0.00$ & $8.32 \pm 0.05$ & $0.86 \pm 0.00$ \\
\end{tabular}
\end{center}
\end{table}

\begin{table}[t]
\caption{AL strategy AUAC and final-step test accuracy on MNIST dataset with ResNet18 classifier, 1000 acquisitions per step, and 1000 initial labels. Averages and standard deviations are computed over three seeds. The method with the highest mean performance is bolded, as well as any method whose $1$ standard deviation bands include that mean. Test accuracy columns are not bolded, as nearly all methods overlap in performance.}
\label{tab:mnist_resnet}
\begin{center}
\begin{tabular}{p{1.7cm} p{1.7cm} p{1.7cm} p{1.7cm} p{1.7cm} p{1.7cm} p{1.7cm}}
\multicolumn{1}{c}{\bf Strategy}  & \multicolumn{2}{c}{\bf Balanced} & \multicolumn{2}{c}{\bf Imbalanced} & \multicolumn{2}{c}{\bf Imbalanced weighted} \\
& AUAC & Test acc. & AUAC & Test acc. & AUAC & Test acc.
\\ \hline \\
 \textsc{Entropy} & $\mathbf{9.92} \pm 0.00$ & $1.00 \pm 0.00$ & $\mathbf{9.83} \pm 0.03$ & $0.99 \pm 0.00$ & $\mathbf{9.86} \pm 0.02$ & $0.99 \pm 0.00$ \\
 \textsc{Margin} & $\mathbf{9.92} \pm 0.00$ & $1.00 \pm 0.00$ & $\mathbf{9.84} \pm 0.03$ & $0.99 \pm 0.00$ & $\mathbf{9.86} \pm 0.02$ & $0.99 \pm 0.00$ \\
 \textsc{LstConf} & $\mathbf{9.92} \pm 0.00$ & $1.00 \pm 0.00$ & $\mathbf{9.83} \pm 0.03$ & $0.99 \pm 0.00$ & $\mathbf{9.86} \pm 0.02$ & $0.99 \pm 0.00$ \\
 \textsc{KCGrdy} & $9.88 \pm 0.00$ & $0.99 \pm 0.00$ & $9.74 \pm 0.04$ & $0.98 \pm 0.00$ & $9.78 \pm 0.02$ & $0.99 \pm 0.00$ \\
 \textsc{LLoss} & $9.87 \pm 0.00$ & $0.99 \pm 0.00$ & $9.71 \pm 0.03$ & $0.98 \pm 0.01$ & $9.74 \pm 0.02$ & $0.99 \pm 0.00$ \\
 \textsc{VAAL} & $9.88 \pm 0.00$ & $0.99 \pm 0.00$ & $9.74 \pm 0.03$ & $0.99 \pm 0.00$ & $9.80 \pm 0.03$ & $0.99 \pm 0.00$ \\
 \textsc{UncGCN} & $9.88 \pm 0.00$ & $0.99 \pm 0.00$ & $9.75 \pm 0.05$ & $0.99 \pm 0.00$ & $9.79 \pm 0.03$ & $0.99 \pm 0.00$ \\
 \textsc{CoreGCN} & $9.88 \pm 0.00$ & $0.99 \pm 0.00$ & $9.75 \pm 0.03$ & $0.98 \pm 0.00$ & $9.79 \pm 0.03$ & $0.99 \pm 0.00$ \\
 \textsc{HALUni} & $9.85 \pm 0.01$ & $0.99 \pm 0.00$ & $9.68 \pm 0.05$ & $0.98 \pm 0.00$ & $9.76 \pm 0.01$ & $0.99 \pm 0.00$ \\
 \textsc{HALGau} & $9.74 \pm 0.01$ & $0.98 \pm 0.00$ & $9.51 \pm 0.04$ & $0.97 \pm 0.00$ & $9.67 \pm 0.03$ & $0.98 \pm 0.00$ \\
 \textsc{CBAL} & $9.88 \pm 0.01$ & $0.99 \pm 0.00$ & $9.75 \pm 0.02$ & $0.99 \pm 0.00$ & $9.80 \pm 0.02$ & $0.99 \pm 0.00$ \\
 \textsc{Random} & $9.88 \pm 0.00$ & $0.99 \pm 0.00$ & $9.74 \pm 0.04$ & $0.99 \pm 0.00$ & $9.80 \pm 0.02$ & $0.99 \pm 0.00$ \\
\end{tabular}
\end{center}
\end{table}

\subsection{Oracle experiments} \label{app:or_exp}
We refer to section~\ref{app:np_exp} for additional \textsc{Oracle} results.

\subsection{Neural Process experiments} \label{app:np_exp}
In this section we provide additional Neural Process results.

\subsubsection{Additional results: Logistic regression.} 
\begin{figure}[ht] 
\begin{center}
\includegraphics[width=\textwidth]{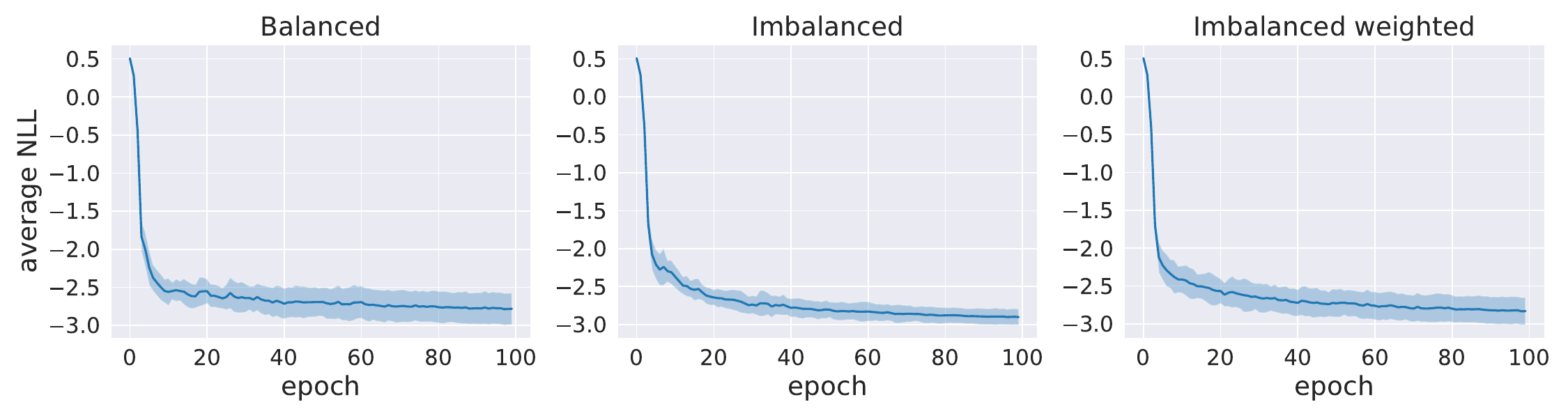}
\end{center}
\caption{Learning curves -- negative log likelihood (NLL) --for the AttnCNP model for the tenth acquisition step on the waveform dataset with logistic regression classifier. Standard deviation computed over nine seeds.}
\label{fig:waveform_logistic_nplc}
\end{figure}

In this section we provide additional results for our logistic regression experiments. First, Figure~\ref{fig:waveform_logistic_nplc} shows average learning curves for the AttnCNP model for the tenth acquisition round of the waveform dataset on all three settings. The negative log likelihood (NLL) of the AttnCNP smoothly decreases, suggesting stable learning of the model. Qualitatively similar results are observed for the other acquisition rounds.

Table~\ref{tab:waveform_logistic_prs_rare} shows precision and recall on the rare class for the waveform dataset. Interestingly, we note that precision is favoured by the classifier in the Imbalanced setting, while recall is favoured in the Imbalanced weighted setting.

\begin{table}[t]
\caption{Final-step Precision and Recall per AL strategy for the rare class on the UCI waveform dataset with a logistic regression classifier. 1 acquisition per step, and 100 initial labels. Averages and standard deviations are computed over nine seeds.}
\label{tab:waveform_logistic_prs_rare}
\begin{center}
\begin{tabular}{p{1.7cm} p{1.7cm} p{1.7cm} p{1.7cm} p{1.7cm} p{1.7cm} p{1.7cm}}
\multicolumn{1}{c}{\bf Strategy} & \multicolumn{2}{c}{\bf Balanced} & \multicolumn{2}{c}{\bf Imbalanced} & \multicolumn{2}{c}{\bf Imbalanced weighted} \\ 
& Precision & Recall & Precision & Recall & Precision & Recall
\\ \hline \\
 \textsc{Oracle} & $0.95 \pm 0.02$ & $0.90 \pm 0.03$ & $0.98 \pm 0.03$ & $0.78 \pm 0.09$ & $0.70 \pm 0.14$ & $0.91 \pm 0.06$ \\
\\ \hdashline \\
 \textsc{UncSamp} & $0.88 \pm 0.02$ & $0.86 \pm 0.04$ & $0.95 \pm 0.07$ & $0.70 \pm 0.09$ & $0.70 \pm 0.16$ & $0.77 \pm 0.06$ \\
 \textsc{KCGrdy} & $0.88 \pm 0.03$ & $0.86 \pm 0.04$ & $0.94 \pm 0.07$ & $0.68 \pm 0.08$ & $0.71 \pm 0.18$ & $0.77 \pm 0.09$ \\
 \textsc{HALUni} & $0.88 \pm 0.04$ & $0.85 \pm 0.04$ & $0.94 \pm 0.08$ & $0.62 \pm 0.11$ & $0.66 \pm 0.20$ & $0.75 \pm 0.10$ \\
 \textsc{HALGau} & $0.90 \pm 0.02$ & $0.84 \pm 0.04$ & $0.94 \pm 0.07$ & $0.64 \pm 0.11$ & $0.69 \pm 0.18$ & $0.73 \pm 0.10$ \\
 \textsc{CBAL} & $0.88 \pm 0.04$ & $0.86 \pm 0.04$ & $0.95 \pm 0.06$ & $0.69 \pm 0.08$ & $0.66 \pm 0.13$ & $0.78 \pm 0.07$ \\
 \textsc{Random} & $0.89 \pm 0.03$ & $0.84 \pm 0.05$ & $0.95 \pm 0.07$ & $0.65 \pm 0.10$ & $0.66 \pm 0.17$ & $0.74 \pm 0.10$ \\
 \textsc{NP} & $0.89 \pm 0.03$ & $0.85 \pm 0.04$ & $0.93 \pm 0.08$ & $0.67 \pm 0.11$ & $0.71 \pm 0.18$ & $0.78 \pm 0.08$ \\
\end{tabular}
\end{center}
\end{table}

\begin{figure}[ht] 
\begin{center}
\includegraphics[width=\textwidth]{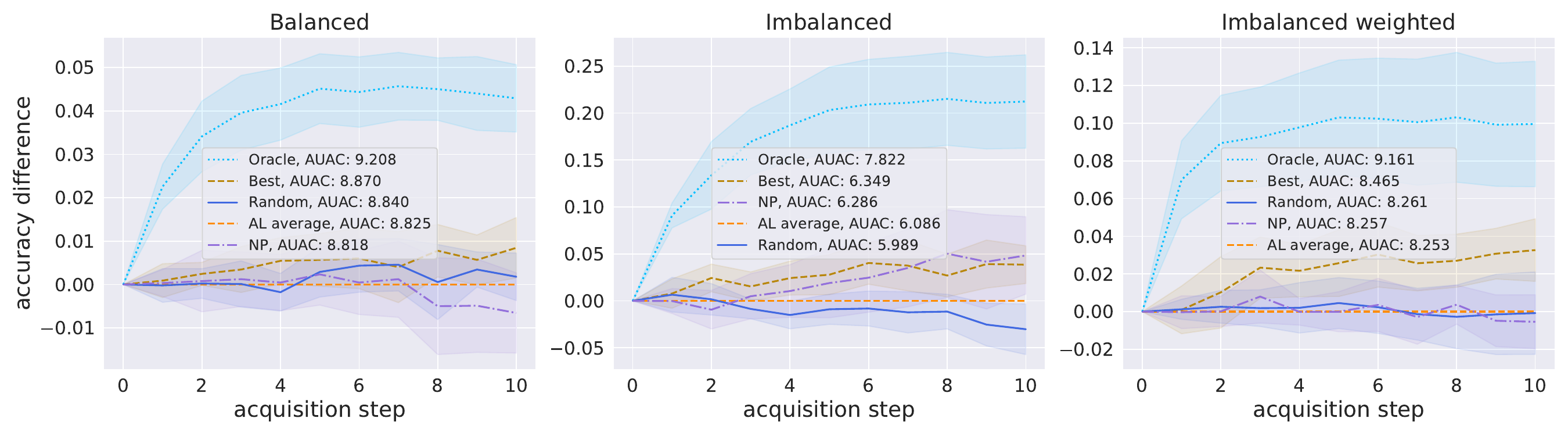}
\end{center}
\caption{Relative performance of acquisition strategies for mushrooms dataset and logistic regression classifier, 1 acquisition per step, and 100 initial labels. Accuracy differences of \textsc{Random}, \textsc{Oracle}, \textsc{NP} and the best remaining AL strategy (\textsc{Best}) are computed w.r.t. the average of remaining AL strategies (\textsc{AL average}). Shaded region represents twice the standard error of the mean over nine seeds.}
\label{fig:mushrooms_logistic}
\end{figure}

\begin{figure}[ht] 
\begin{center}
\includegraphics[width=\textwidth]{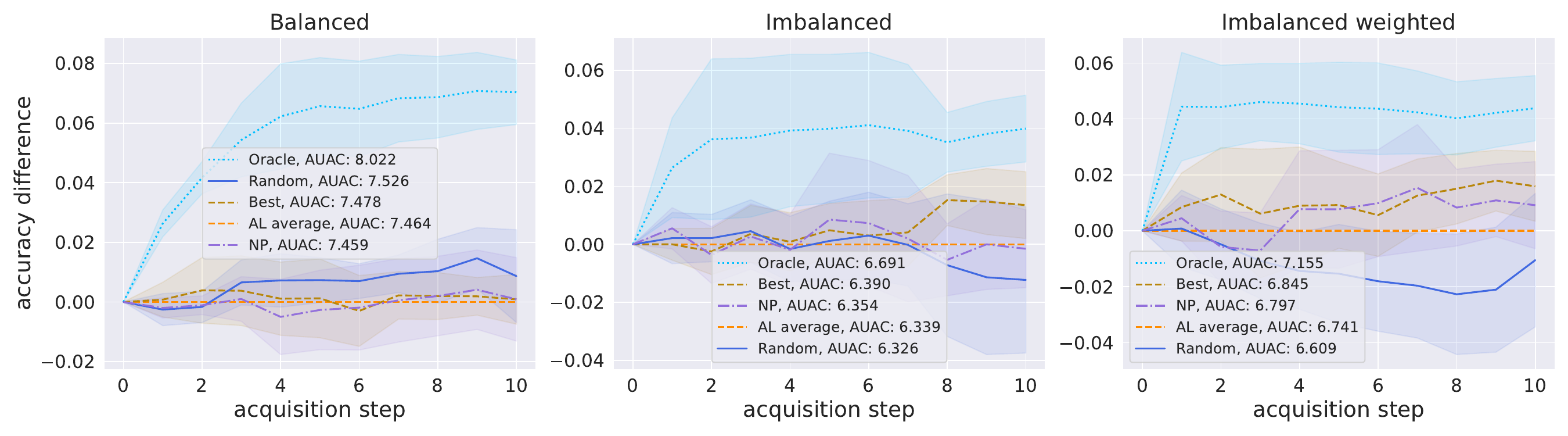}
\end{center}
\caption{Relative performance of acquisition strategies for adult dataset and logistic regression classifier, 1 acquisition per step, and 100 initial labels. Accuracy differences of \textsc{Random}, \textsc{Oracle}, \textsc{NP} and the best remaining AL strategy (\textsc{Best}) are computed w.r.t. the average of remaining AL strategies (\textsc{AL average}). Shaded region represents twice the standard error of the mean over nine seeds.}
\label{fig:adult_logistic}
\end{figure}

Table~\ref{tab:mushrooms_logistic} and~\ref{tab:adult_logistic} show all remaining accuracy and AUAC results on respectively the mushrooms and adult dataset. Figures~\ref{fig:mushrooms_logistic} and~\ref{fig:adult_logistic} show the performance as a function of acquisition step.

\begin{table}[t]
\caption{AL strategy AUAC and final-step test accuracy on UCI mushrooms dataset with logistic regression classifier, 1 acquisition per step, and 100 initial labels. Averages and standard deviations are computed over nine seeds.}
\label{tab:mushrooms_logistic}
\begin{center}
\begin{tabular}{p{1.7cm} p{1.7cm} p{1.7cm} p{1.7cm} p{1.7cm} p{1.7cm} p{1.7cm}}
\multicolumn{1}{c}{\bf Strategy}  & \multicolumn{2}{c}{\bf Balanced} & \multicolumn{2}{c}{\bf Imbalanced} & \multicolumn{2}{c}{\bf Imbalanced weighted} \\
& AUAC & Test acc. & AUAC & Test acc. & AUAC & Test acc.
\\ \hline \\
 \textsc{Oracle} & $\mathbf{9.21} \pm 0.14$ & $\mathbf{0.93} \pm 0.01$ & $\mathbf{7.82} \pm 0.75$ & $\mathbf{0.83} \pm 0.09$ & $\mathbf{9.16} \pm 0.25$ & $\mathbf{0.93} \pm 0.02$ \\
 \\ \hdashline \\
 \textsc{UncSamp} & $8.87 \pm 0.15$ & $0.90 \pm 0.01$ & $6.35 \pm 0.62$ & $0.66 \pm 0.06$ & $8.18 \pm 0.66$ & $0.83 \pm 0.07$ \\
 \textsc{KCGrdy} & $8.82 \pm 0.14$ & $0.89 \pm 0.02$ & $6.20 \pm 0.39$ & $0.63 \pm 0.06$ & $8.46 \pm 0.54$ & $0.86 \pm 0.04$ \\
 \textsc{HALUni} & $8.81 \pm 0.20$ & $0.88 \pm 0.02$ & $5.91 \pm 0.54$ & $0.59 \pm 0.05$ & $8.22 \pm 0.63$ & $0.83 \pm 0.06$ \\
 \textsc{HALGau} & $8.78 \pm 0.18$ & $0.88 \pm 0.02$ & $5.90 \pm 0.54$ & $0.59 \pm 0.05$ & $8.24 \pm 0.62$ & $0.83 \pm 0.06$ \\
 \textsc{CBAL} & $8.85 \pm 0.15$ & $0.89 \pm 0.01$ & $6.07 \pm 0.71$ & $0.62 \pm 0.07$ & $8.15 \pm 0.64$ & $0.81 \pm 0.07$ \\
 \textsc{Random} & $8.84 \pm 0.17$ & $0.89 \pm 0.01$ & $5.99 \pm 0.52$ & $0.59 \pm 0.04$ & $8.26 \pm 0.61$ & $0.83 \pm 0.07$ \\
 \textsc{NP} & $8.82 \pm 0.22$ & $0.88 \pm 0.03$ & $6.29 \pm 0.55$ & $0.67 \pm 0.06$ & $8.26 \pm 0.56$ & $0.83 \pm 0.05$ \\
\end{tabular}
\end{center}
\end{table}

\begin{table}[t]
\caption{AL strategy AUAC and final-step test accuracy on UCI adult dataset with logistic regression classifier, 1 acquisition per step, and 100 initial labels. Averages and standard deviations are computed over nine seeds.}
\label{tab:adult_logistic}
\begin{center}
\begin{tabular}{p{1.7cm} p{1.7cm} p{1.7cm} p{1.7cm} p{1.7cm} p{1.7cm} p{1.7cm}}
\multicolumn{1}{c}{\bf Strategy}  & \multicolumn{2}{c}{\bf Balanced} & \multicolumn{2}{c}{\bf Imbalanced} & \multicolumn{2}{c}{\bf Imbalanced weighted} \\
& AUAC & Test acc. & AUAC & Test acc. & AUAC & Test acc.
\\ \hline \\
 \textsc{Oracle} & $\mathbf{8.02} \pm 0.37$ & $\mathbf{0.82} \pm 0.04$ & $\mathbf{6.69} \pm 0.59$ & $\mathbf{0.68} \pm 0.05$ & $\mathbf{7.16} \pm 0.63$ & $\mathbf{0.72} \pm 0.06$ \\
 \\ \hdashline \\
 \textsc{UncSamp} & $7.46 \pm 0.43$ & $0.75 \pm 0.04$ & $6.38 \pm 0.41$ & $0.66 \pm 0.05$ & $6.85 \pm 0.50$ & $0.69 \pm 0.06$ \\
 \textsc{KCGrdy} & $7.47 \pm 0.40$ & $0.74 \pm 0.04$ & $6.38 \pm 0.52$ & $0.65 \pm 0.06$ & $6.77 \pm 0.66$ & $0.68 \pm 0.07$ \\
 \textsc{HALUni} & $7.48 \pm 0.49$ & $0.75 \pm 0.04$ & $6.30 \pm 0.47$ & $0.63 \pm 0.04$ & $6.66 \pm 0.55$ & $0.66 \pm 0.06$ \\
 \textsc{HALGau} & $7.44 \pm 0.45$ & $0.74 \pm 0.05$ & $6.26 \pm 0.47$ & $0.62 \pm 0.05$ & $6.64 \pm 0.55$ & $0.66 \pm 0.05$ \\
 \textsc{CBAL} & $7.47 \pm 0.42$ & $0.74 \pm 0.04$ & $6.39 \pm 0.42$ & $0.66 \pm 0.05$ & $6.79 \pm 0.55$ & $0.69 \pm 0.07$ \\
 \textsc{Random} & $7.53 \pm 0.41$ & $0.75 \pm 0.05$ & $6.33 \pm 0.52$ & $0.63 \pm 0.05$ & $6.61 \pm 0.59$ & $0.67 \pm 0.06$ \\
 \textsc{NP} & $7.46 \pm 0.47$ & $0.75 \pm 0.05$ & $6.35 \pm 0.50$ & $0.64 \pm 0.06$ & $6.80 \pm 0.58$ & $0.69 \pm 0.07$ \\
\end{tabular}
\end{center}
\end{table}

\subsubsection{Additional results: SVM.}
\begin{table}[t]
\caption{AL strategy AUAC and final-step test accuracy on UCI waveform dataset with SVM classifier, 1 acquisition per step, and 100 initial labels. Averages and standard deviations are computed over nine seeds.}
\label{tab:waveform_svm}
\begin{center}
\begin{tabular}{p{1.7cm} p{1.7cm} p{1.7cm} p{1.7cm} p{1.7cm} p{1.7cm} p{1.7cm}}
\multicolumn{1}{c}{\bf Strategy}  & \multicolumn{2}{c}{\bf Balanced} & \multicolumn{2}{c}{\bf Imbalanced} & \multicolumn{2}{c}{\bf Imbalanced weighted} \\
& AUAC & Test acc. & AUAC & Test acc. & AUAC & Test acc.
\\ \hline \\
 \textsc{Oracle} & $\mathbf{9.22} \pm 0.10$ & $\mathbf{0.93} \pm 0.01$ & $\mathbf{8.26} \pm 0.46$ & $0.84 \pm 0.05$ & $\mathbf{9.30} \pm 0.26$ & $\mathbf{0.95} \pm 0.02$ \\
 \\ \hdashline \\
 \textsc{FScore} & $8.87 \pm 0.15$ & $0.89 \pm 0.02$ & $8.10 \pm 0.53$ & $\mathbf{0.85} \pm 0.05$ & $8.71 \pm 0.47$ & $0.88 \pm 0.06$ \\
 \textsc{KCGrdy} & $8.85 \pm 0.14$ & $0.88 \pm 0.01$ & $7.93 \pm 0.56$ & $0.81 \pm 0.06$ & $8.57 \pm 0.45$ & $0.87 \pm 0.03$ \\
 \textsc{Random} & $8.86 \pm 0.16$ & $0.89 \pm 0.02$ & $7.47 \pm 0.53$ & $0.75 \pm 0.06$ & $8.54 \pm 0.51$ & $0.86 \pm 0.05$ \\
 \textsc{NP} & $8.85 \pm 0.16$ & $0.89 \pm 0.02$ & $7.75 \pm 0.60$ & $0.80 \pm 0.07$ & $8.50 \pm 0.61$ & $0.86 \pm 0.06$ \\
\end{tabular}
\end{center}
\end{table}

\begin{figure}[ht] 
\begin{center}
\includegraphics[width=\textwidth]{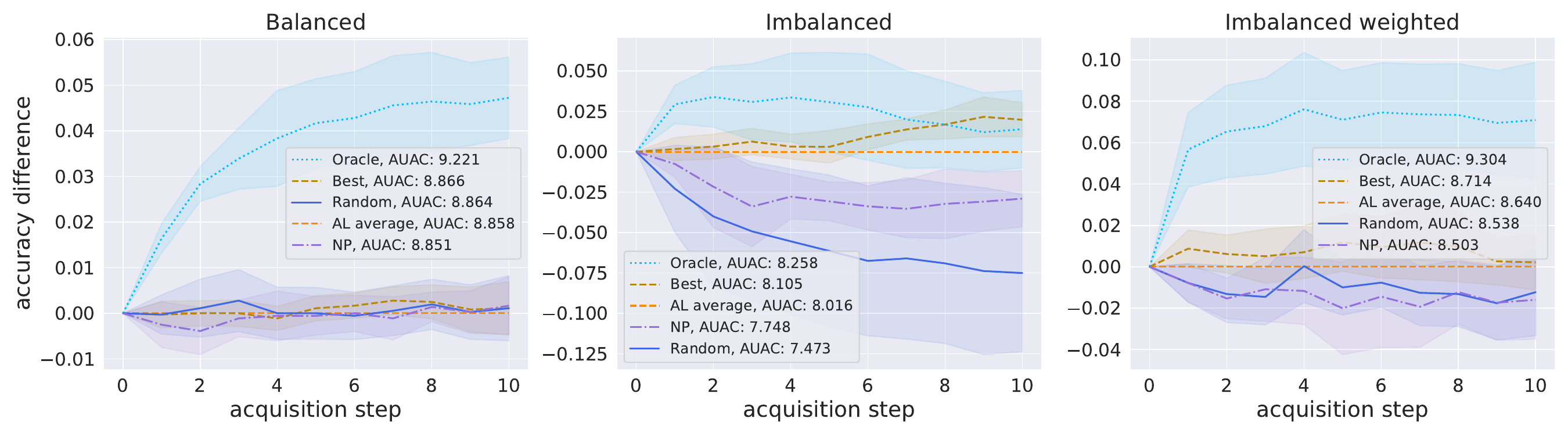}
\end{center}
\caption{Relative performance of acquisition strategies for waveform dataset and SVM classifier, 1 acquisition per step, and 100 initial labels. Accuracy differences of \textsc{Random}, \textsc{Oracle}, \textsc{NP} and the best remaining AL strategy (\textsc{Best}) are computed w.r.t. the average of remaining AL strategies (\textsc{AL average}). Shaded region represents twice the standard error of the mean over nine seeds.}
\label{fig:waveform_svm}
\end{figure}

\begin{figure}[ht] 
\begin{center}
\includegraphics[width=\textwidth]{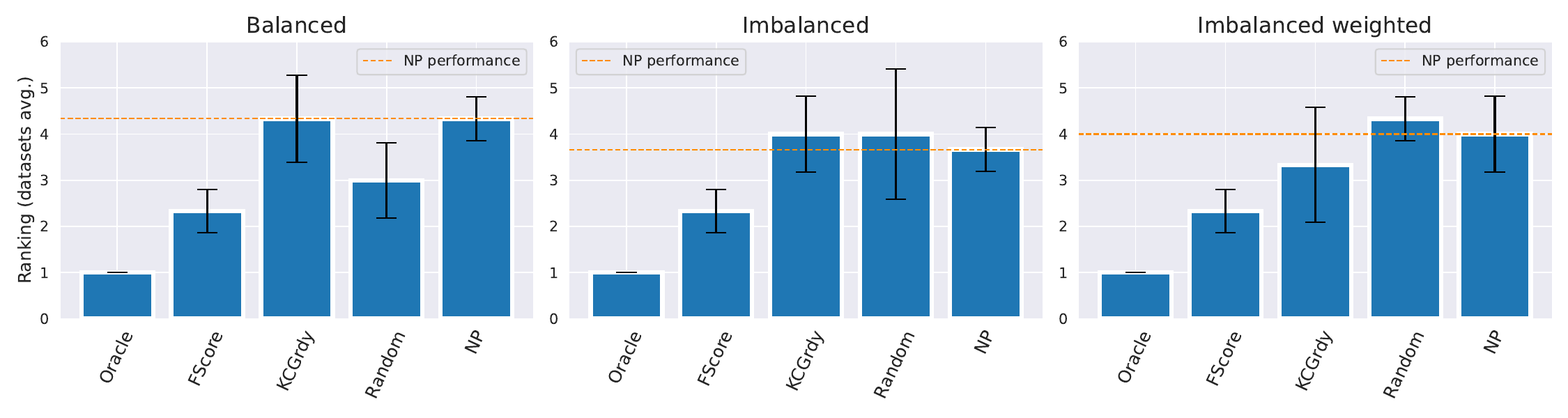}
\end{center}
\caption{Relative AUAC rank of AL strategies averaged over the three UCI datasets for SVM. Standard deviation of this rank is denoted by the error bars.}
\label{fig:rank_svm_AUAC_uci}
\end{figure}

\begin{figure}[ht] 
\begin{center}
\includegraphics[width=\textwidth]{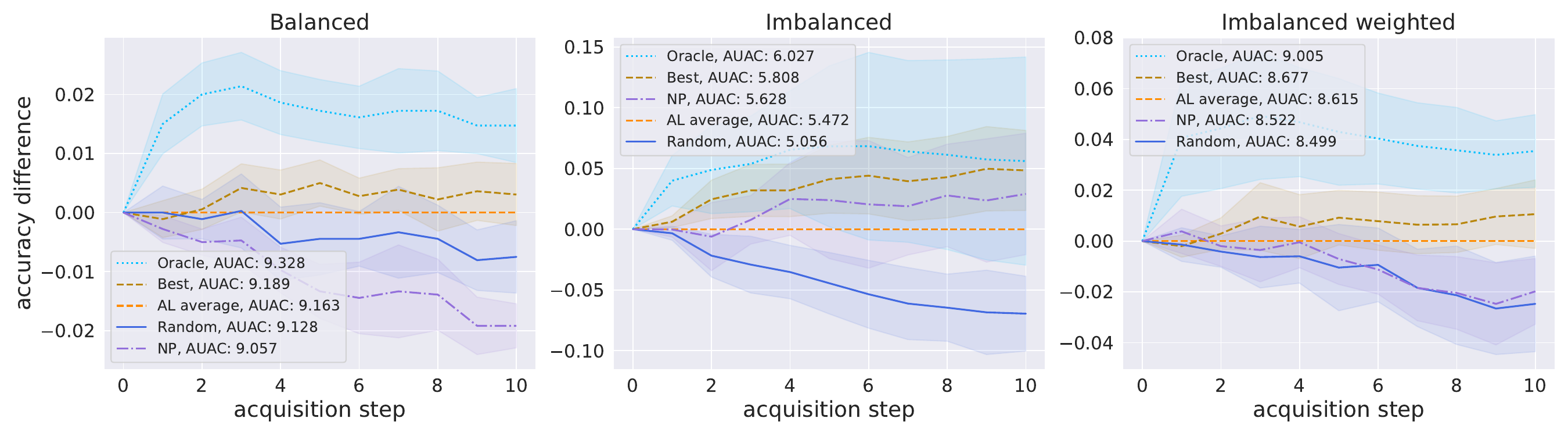}
\end{center}
\caption{Relative performance of acquisition strategies for mushrooms dataset and SVM classifier, 1 acquisition per step, and 100 initial labels. Accuracy differences of \textsc{Random}, \textsc{Oracle}, \textsc{NP} and the best remaining AL strategy (\textsc{Best}) are computed w.r.t. the average of remaining AL strategies (\textsc{AL average}). Shaded region represents twice the standard error of the mean over nine seeds.}
\label{fig:mushrooms_svm}
\end{figure}

\begin{figure}[ht] 
\begin{center}
\includegraphics[width=\textwidth]{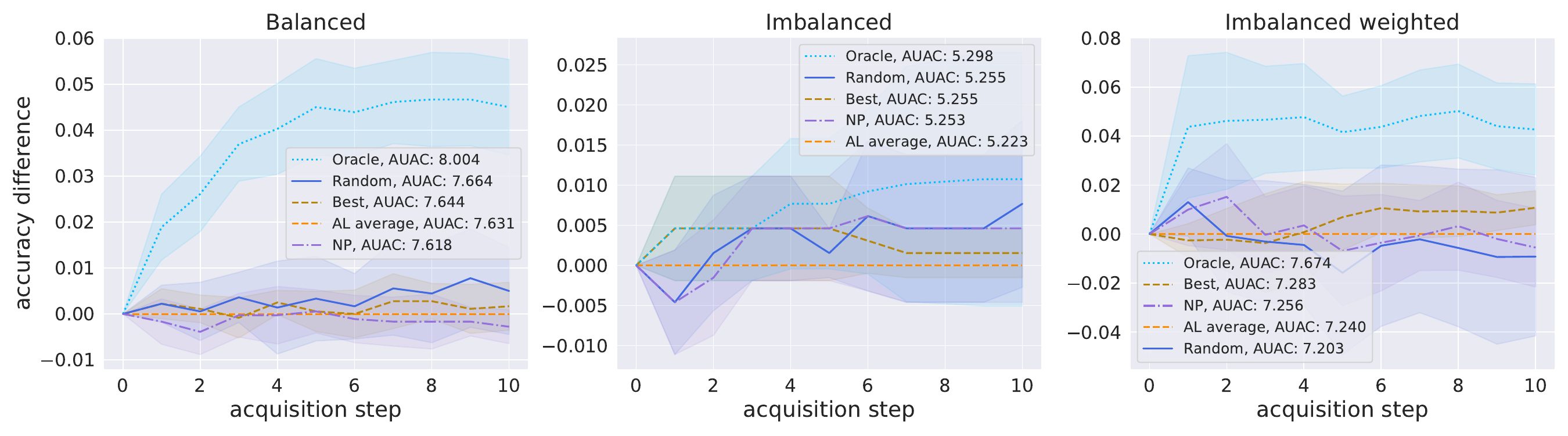}
\end{center}
\caption{Relative performance of acquisition strategies for adult dataset and SVM classifier, 1 acquisition per step, and 100 initial labels. Accuracy differences of \textsc{Random}, \textsc{Oracle}, \textsc{NP} and the best remaining AL strategy (\textsc{Best}) are computed w.r.t. the average of remaining AL strategies (\textsc{AL average}). Shaded region represents twice the standard error of the mean over nine seeds.}
\label{fig:adult_svm}
\end{figure}

Here we provide additional results for an SVM classifier. Table~\ref{tab:waveform_svm} and Figure~\ref{fig:waveform_svm} show performance on the waveform dataset with the SVM classifier. Table~\ref{tab:mushrooms_svm}, Figure~\ref{fig:mushrooms_svm} and Table~\ref{tab:adult_svm}, Figure~\ref{fig:adult_svm} show similar results for respectively the mushrooms and adult dataset. As noted previously, \textsc{FScore} consistently ranks highest of the AL methods. Since \textsc{AL average} consists of only \textsc{FScore} and \textsc{KCGrdy} here, its strong performance is primarily driven by \textsc{FScore}. Even so, this setting seems more difficult for \textsc{NP} to learn, suggesting that the choice of underlying classifier is important. Figure~\ref{fig:rank_svm_AUAC_uci} corroborates this hypothesis, although it again suggests that \textsc{NP} is specifically more promising for imbalanced objectives.

\begin{figure}[ht] 
\begin{center}
\includegraphics[width=\textwidth]{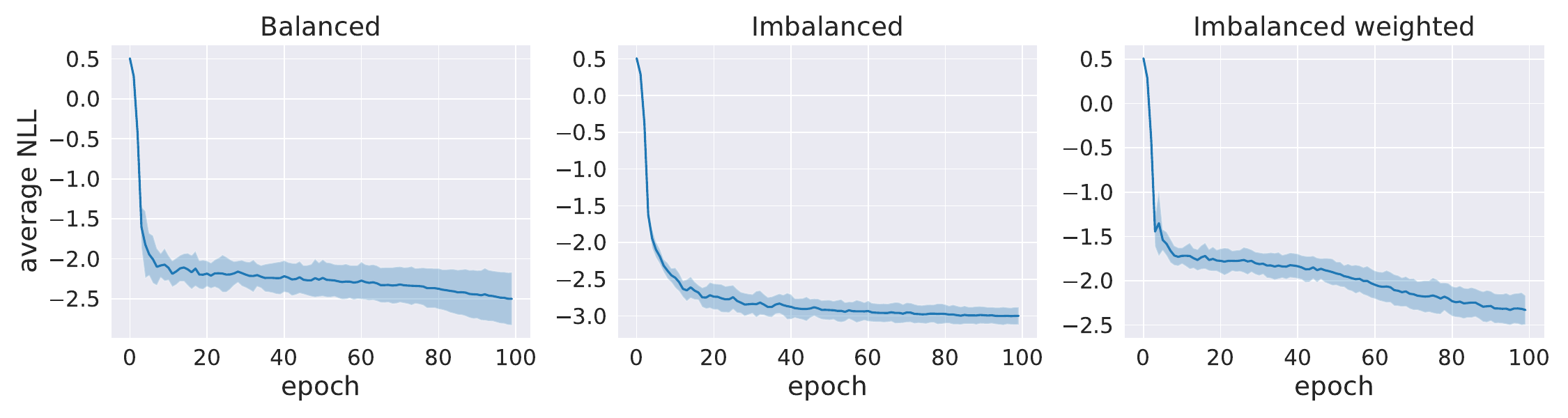}
\end{center}
\caption{Learning curves -- negative log likelihood (NLL) --for the AttnCNP model for the tenth acquisition step on the waveform dataset with SVM classifier. Standard deviation computed over nine seeds.}
\label{fig:waveform_svm_nplc}
\end{figure}

Figure~\ref{fig:waveform_svm_nplc} shows average learning curves for the AttnCNP model for the tenth acquisition round of the waveform dataset with SVM classifier on all three settings. The negative log likelihood (NLL) of the AttnCNP smoothly decreases, suggesting stable learning of the model. Qualitatively similar results are observed for the other acquisition rounds.

Table~\ref{tab:waveform_svm_prs_rare} shows precision and recall on the rare class for the waveform dataset with SVM classifier. Interestingly, we note that precision is favoured by the classifier in both imbalanced settings, except for the Oracle strategy, which favours recall for Imbalanced weighted.

\begin{table}[t]
\caption{Final-step Precision and Recall per AL strategy for the rare class on the UCI waveform dataset with an SVM classifier. 1 acquisition per step, and 100 initial labels. Averages and standard deviations are computed over nine seeds.}
\label{tab:waveform_svm_prs_rare}
\begin{center}
\begin{tabular}{p{1.7cm} p{1.7cm} p{1.7cm} p{1.7cm} p{1.7cm} p{1.7cm} p{1.7cm}}
\multicolumn{1}{c}{\bf Strategy} & \multicolumn{2}{c}{\bf Balanced} & \multicolumn{2}{c}{\bf Imbalanced} & \multicolumn{2}{c}{\bf Imbalanced weighted} \\ 
& Precision & Recall & Precision & Recall & Precision & Recall
\\ \hline \\
 \textsc{Oracle} & $0.98 \pm 0.01$ & $0.89 \pm 0.03$ & $0.99 \pm 0.02$ & $0.68 \pm 0.10$ & $0.86 \pm 0.11$ & $0.91 \pm 0.05$ \\
\\ \hdashline \\
 \textsc{FScore} & $0.92 \pm 0.04$ & $0.86 \pm 0.04$ & $0.98 \pm 0.03$ & $0.69 \pm 0.10$ & $0.80 \pm 0.06$ & $0.77 \pm 0.12$ \\
 \textsc{KCGrdy} & $0.94 \pm 0.02$ & $0.82 \pm 0.04$ & $0.99 \pm 0.03$ & $0.61 \pm 0.11$ & $0.81 \pm 0.20$ & $0.77 \pm 0.09$ \\
 \textsc{Random} & $0.95 \pm 0.03$ & $0.82 \pm 0.04$ & $0.99 \pm 0.03$ & $0.50 \pm 0.11$ & $0.89 \pm 0.12$ & $0.73 \pm 0.11$ \\
 \textsc{NP} & $0.94 \pm 0.03$ & $0.83 \pm 0.04$ & $0.99 \pm 0.03$ & $0.59 \pm 0.13$ & $0.81 \pm 0.18$ & $0.74 \pm 0.13$ \\
\end{tabular}
\end{center}
\end{table}

\begin{figure}[ht] 
\begin{center}
\includegraphics[width=\textwidth]{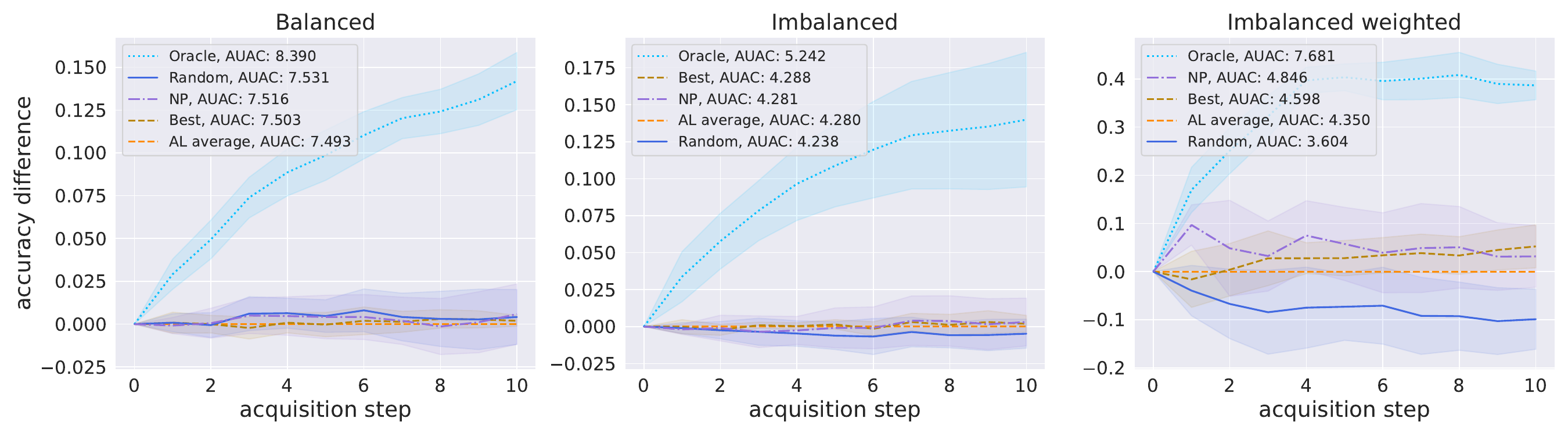}
\end{center}
\caption{Relative performance of acquisition strategies for MNIST dataset and SVM classifier, 1 acquisition per step, and 100 initial labels. Accuracy differences of \textsc{Random}, \textsc{Oracle}, \textsc{NP} and the best remaining AL strategy (\textsc{Best}) are computed w.r.t. the average of remaining AL strategies (\textsc{AL average}). Shaded region represents twice the standard error of the mean over nine seeds.}
\label{fig:mnist_svm}
\end{figure}

We provide multiclass MNIST SVM experiments, shown in Table~\ref{tab:mnist_svm} and Figure~\ref{fig:mnist_svm}. The naive \textsc{FScore} strategy -- finding the datapoint closest to any of the ten one-vs-rest decision hyperplanes -- experiences a strong drop in performance, indicating that less naive generalisations such as those in \cite{svm2014} are necessary.

\begin{table}[t]
\caption{AL strategy AUAC and final-step test accuracy on UCI mushrooms dataset with SVM classifier, 1 acquisition per step, and 100 initial labels. Averages and standard deviations are computed over nine seeds.}
\label{tab:mushrooms_svm}
\begin{center}
\begin{tabular}{p{1.7cm} p{1.7cm} p{1.7cm} p{1.7cm} p{1.7cm} p{1.7cm} p{1.7cm}}
\multicolumn{1}{c}{\bf Strategy}  & \multicolumn{2}{c}{\bf Balanced} & \multicolumn{2}{c}{\bf Imbalanced} & \multicolumn{2}{c}{\bf Imbalanced weighted} \\
& AUAC & Test acc. & AUAC & Test acc. & AUAC & Test acc.
\\ \hline \\
 \textsc{Oracle} & $\mathbf{9.33} \pm 0.16$ & $\mathbf{0.94} \pm 0.02$ & $\mathbf{6.03} \pm 0.87$ & $\mathbf{0.63} \pm 0.10$ & $\mathbf{9.00} \pm 0.18$ & $\mathbf{0.91} \pm 0.01$ \\
 \\ \hdashline \\
 \textsc{FScore} & $9.19 \pm 0.13$ & $0.93 \pm 0.02$ & $5.81 \pm 0.53$ & $0.62 \pm 0.08$ & $8.55 \pm 0.44$ & $0.86 \pm 0.04$ \\
 \textsc{KCGrdy} & $9.14 \pm 0.19$ & $0.92 \pm 0.02$ & $5.14 \pm 0.14$ & $0.53 \pm 0.03$ & $8.68 \pm 0.29$ & $0.88 \pm 0.02$ \\
 \textsc{Random} & $9.13 \pm 0.15$ & $0.92 \pm 0.01$ & $5.06 \pm 0.08$ & $0.51 \pm 0.01$ & $8.50 \pm 0.48$ & $0.85 \pm 0.04$ \\
 \textsc{NP} & $9.06 \pm 0.20$ & $0.90 \pm 0.02$ & $5.63 \pm 0.51$ & $0.60 \pm 0.06$ & $8.52 \pm 0.43$ & $0.85 \pm 0.04$ \\
\end{tabular}
\end{center}
\end{table}

\begin{table}[t]
\caption{AL strategy AUAC and final-step test accuracy on UCI adult dataset with SVM classifier, 1 acquisition per step, and 100 initial labels. Averages and standard deviations are computed over nine seeds.}
\label{tab:adult_svm}
\begin{center}
\begin{tabular}{p{1.7cm} p{1.7cm} p{1.7cm} p{1.7cm} p{1.7cm} p{1.7cm} p{1.7cm}}
\multicolumn{1}{c}{\bf Strategy}  & \multicolumn{2}{c}{\bf Balanced} & \multicolumn{2}{c}{\bf Imbalanced} & \multicolumn{2}{c}{\bf Imbalanced weighted} \\
& AUAC & Test acc. & AUAC & Test acc. & AUAC & Test acc.
\\ \hline \\
 \textsc{Oracle} & $\mathbf{8.00} \pm 0.38$ & $\mathbf{0.81} \pm 0.04$ & $\mathbf{5.30} \pm 0.37$ & $\mathbf{0.53} \pm 0.04$ & $\mathbf{7.67} \pm 0.54$ & $\mathbf{0.78} \pm 0.05$ \\
 \\ \hdashline \\
 \textsc{FScore} & $7.64 \pm 0.41$ & $0.77 \pm 0.04$ & $5.25 \pm 0.36$ & $0.52 \pm 0.04$ & $7.28 \pm 0.51$ & $0.74 \pm 0.05$ \\
 \textsc{KCGrdy} & $7.62 \pm 0.41$ & $0.76 \pm 0.04$ & $5.19 \pm 0.32$ & $0.52 \pm 0.04$ & $7.20 \pm 0.67$ & $0.72 \pm 0.07$ \\
 \textsc{Random} & $7.66 \pm 0.43$ & $0.77 \pm 0.04$ & $5.25 \pm 0.35$ & $0.53 \pm 0.04$ & $7.20 \pm 0.60$ & $0.72 \pm 0.06$ \\
 \textsc{NP} & $7.62 \pm 0.41$ & $0.76 \pm 0.04$ & $5.25 \pm 0.36$ & $0.53 \pm 0.04$ & $7.26 \pm 0.61$ & $0.73 \pm 0.06$ \\
\end{tabular}
\end{center}
\end{table}

\begin{table}[t]
\caption{AL strategy AUAC and final-step test accuracy on MNIST dataset with SVM classifier, 1 acquisition per step, and 100 initial labels. Averages and standard deviations are computed over nine seeds.}
\label{tab:mnist_svm}
\begin{center}
\begin{tabular}{p{1.7cm} p{1.7cm} p{1.7cm} p{1.7cm} p{1.7cm} p{1.7cm} p{1.7cm}}
\multicolumn{1}{c}{\bf Strategy}  & \multicolumn{2}{c}{\bf Balanced} & \multicolumn{2}{c}{\bf Imbalanced} & \multicolumn{2}{c}{\bf Imbalanced weighted} \\
& AUAC & Test acc. & AUAC & Test acc. & AUAC & Test acc.
\\ \hline \\
 \textsc{Oracle} & $\mathbf{8.39} \pm 0.15$ & $\mathbf{0.89} \pm 0.01$ & $\mathbf{5.24} \pm 0.50$ & $\mathbf{0.57} \pm 0.07$ & $\mathbf{7.68} \pm 0.40$ & $\mathbf{0.83} \pm 0.05$ \\
 \\ \hdashline \\
 \textsc{FScore} & $7.48 \pm 0.31$ & $0.74 \pm 0.03$ & $4.27 \pm 0.19$ & $0.43 \pm 0.03$ & $4.10 \pm 0.94$ & $0.39 \pm 0.10$ \\
 \textsc{KCGrdy} & $7.50 \pm 0.24$ & $0.75 \pm 0.03$ & $4.29 \pm 0.19$ & $0.43 \pm 0.02$ & $4.60 \pm 0.48$ & $0.49 \pm 0.04$ \\
 \textsc{Random} & $7.53 \pm 0.27$ & $0.75 \pm 0.03$ & $4.24 \pm 0.18$ & $0.43 \pm 0.02$ & $3.60 \pm 0.53$ & $0.34 \pm 0.06$ \\
 \textsc{NP} & $7.52 \pm 0.29$ & $0.75 \pm 0.03$ & $4.28 \pm 0.13$ & $0.43 \pm 0.01$ & $4.85 \pm 0.61$ & $0.47 \pm 0.05$ \\
\end{tabular}
\end{center}
\end{table}

\end{document}